\documentclass[lettersize,journal]{IEEEtran}
\usepackage{amsmath,amsfonts}
\usepackage{algorithmic}
\usepackage{algorithm}
\usepackage{array}
\usepackage[caption=false,font=normalsize,labelfont=sf,textfont=sf]{subfig}
\usepackage{textcomp}
\usepackage{stfloats}
\usepackage{url}
\usepackage{verbatim}
\usepackage{graphicx}
\usepackage{cite}
\usepackage{enumitem}
\usepackage{hyperref}
\hypersetup{
    colorlinks=true, 
    urlcolor=magenta,   
}
\usepackage{indentfirst} 
\usepackage{makecell,multirow,diagbox}
\newtheorem{remark}{Remark}
\usepackage{amsmath,amssymb,amsfonts}
\begin{document}

\title{V-STC: A Time-Efficient Multi-Vehicle Coordinated Trajectory Planning Approach}

\author{Pengfei Liu, Jialing Zhou,~\IEEEmembership{Senior Member, IEEE,} Yuezu Lv,~\IEEEmembership{Senior Member, IEEE,} Guanghui Wen,~\IEEEmembership{Senior Member, IEEE,} and Tingwen Huang,~\IEEEmembership{Fellow, IEEE}
\thanks{Pengfei Liu is with the School of Automation, Beijing Institute of Technology, Beijing 100081, China (e-mail: penngfeiliu@bit.edu.cn).}
\thanks{Jialing Zhou is with the State Key Laboratory of CNS/ATM, Beijing Institute of Technology, Beijing 100081, China, and also with the Beijing Institute of Technology (Zhuhai), Zhuhai 519088, China (e-mail: jlzhou@bit.edu.cn).}
\thanks{Yuezu Lv is with the Beijing Key Laboratory of Lightweight Intelligent System, Beijing Institute of Technology, Beijing 100081, China (e-mail: yzlv@bit.edu.cn).}
\thanks{Guanghui Wen is with the School of Automation, Southeast University, Nanjing 211189, China (e-mail: ghwen@seu.edu.cn).}
\thanks{Tingwen Huang is with the Faculty of Computer Science and Control Engineering, Shenzhen University of Advanced Technology, Shenzhen 518055, China (e-mail: huangtingwen@suat-sz.edu.cn).}
}

\markboth{Journal of \LaTeX\ Class Files,~Vol.~14, No.~8, August~2021}%
{Shell \MakeLowercase{\textit{et al.}}: A Sample Article Using IEEEtran.cls for IEEE Journals}


\maketitle

\begin{abstract}
  Coordinating the motions of multiple autonomous vehicles (AVs) requires planning frameworks that ensure safety while making efficient use of space and time. This paper presents a new approach, termed variable-time-step spatio-temporal corridor (V-STC), that enhances the temporal efficiency of multi-vehicle coordination. An optimization model is formulated to construct a V-STC for each AV, in which both the spatial configuration of the corridor cubes and their time durations are treated as decision variables. By allowing the corridor's spatial position and time step to vary, the constructed V-STC reduces the overall temporal occupancy of each AV while maintaining collision-free separation in the spatio-temporal domain. Based on the generated V-STC, a dynamically feasible trajectory is then planned independently for each AV. Simulation studies demonstrate that the proposed method achieves safe multi-vehicle coordination and yields more time-efficient motion compared with existing STC approaches.
\end{abstract}


\section{Introduction}
\IEEEPARstart{C}{oordinated} trajectory planning for multiple AVs has received increasing attention due to its potential to enhance traffic efficiency and safety in both structured and unstructured environments \cite{Lim2021, Chai2022, Meng2023, Guan2023}. Unlike single-vehicle planners that operate independently, multi-vehicle coordination explicitly leverages interactions among vehicles, enabling more efficient allocation of spatial-temporal resources and improved system-wide performance.

Existing multi-vehicle trajectory planning approaches can be broadly categorized into decoupled planning and joint spatio-temporal optimization. Decoupled methods, such as lateral-longitudinal decomposition \cite{Miller2018, Ding2023} and path-speed decomposition \cite{Zhang2019, Wang2023, Lin2024}, reduce computational complexity by splitting the planning problem into lower-dimensional subproblems. However, the resulting reduction in search space often leads to suboptimal trajectories limited coordination capability. In contrast, optimization-based methods that directly search in the full spatio-temporal domain \cite{Morsali2021, Li2021, Xie2022, Duan2022, Yoon2024} preserve a richer solution space and can generate higher-quality coordinated trajectories, but typically incur substantial computational burden, especially as the number of vehicles grows.

To balance solution quality and computational tractability, recent efforts have introduced spatio-temporal corridor (STC) as a structured representation for multi-vehicle coordination \cite{Zhang2023, Zang2023, Zang2024}. STC-based frameworks typically adopt a two-stage pipeline: (i) constructing a collision-free corridor composed of a sequence of spatio-temporal cubes, and (ii) computing a dynamically feasible trajectory within the corridor. This hierarchical design has demonstrated effectiveness in coordinating multiple vehicles across intersection scenarios and other constrained environments. For instance, \cite{Zang2023} proposed a unified planning and optimization framework based on spatio-temporal safety corridors to ensure both collision avoidance and trajectory smoothness, while \cite{Zang2024} introduced a coordinated behavior and trajectory planning framework for multi-unmanned ground vehicles in narrow unstructured environments through interactive behavior planning and optimization-based trajectory generation.

Despite these advantages, existing STC-based approaches exhibit structural limitations that restrict their flexibility and overall optimality. One limitation is the dependence of corridor construction on predefined reference waypoints. This dependency fixes both the number of corridor cubes and their spatial arrangement, which causes the corridor shape and the resulting trajectory to be heavily influenced by the waypoint layout rather than by the environment or coordination demands. Another limitation arises from the use of fixed temporal durations for corridor cubes. Under such a formulation, the temporal position of each cube is tied to a global time grid while the optimization is applied only to the spatial footprint. As a result, the temporal structure becomes rigid and cannot be adjusted according to AV states or interaction conditions, which reduces temporal coordination flexibility and restricts the full use of available spatio-temporal freedom. These limitations highlight the need for a more adaptable corridor representation that offers spatial flexibility without relying on predefined waypoints and enables temporal allocation responsive to AV motions and interactions rather than uniform time partitioning.

To address these challenges, this paper proposes a V-STC framework for coordinated trajectory planning among multiple AVs. The proposed method constructs a sequence of non-overlapping corridor cubes whose spatial layout and temporal durations are jointly determined by an optimization model, enabling each AV to exploit spatio-temporal freedom more effectively and achieve more time-efficient coordination. After constructing the V-STC, each AV independently generates a smooth and dynamically feasible trajectory within its corridor to guarantee collision-free arrival. The main contributions of this paper are at least twofold:

\begin{enumerate}[leftmargin=*] 
  \item The proposed V-STC provides a waypoint-independent spatial representation. Unlike existing STC-based approaches \cite{Zhang2023, Zang2023, Zang2024} that rely on predefined reference waypoints, the proposed V-STC determines the spatial arrangement of its corridor cubes through optimization. This enables the corridor layout and the required number of cubes to be automatically determined according to feasibility constraints and planning objectives, rather than being fixed by waypoint configuration.

  \item The proposed V-STC incorporates variable time durations for its corridor cubes. By treating the time duration of each cube as an optimization variable, the proposed V-STC reduces the overall spatio-temporal occupancy and enables more flexible temporal coordination among AVs. This leads to more time-efficient multi-vehicle motion compared with fixed-time-step STC formulations \cite{Zhang2023, Zang2023, Zang2024}.
\end{enumerate}

The remainder of this paper is organized as follows. Section \ref{Problem statement and system overview} provides an overview of the proposed framework. Section \ref{Methodology} presents the detailed V-STC construction and trajectory generation algorithm. Section \ref{Experimental Results and Discussion} reports simulation results in multiple scenarios. Section \ref{Conclusion} concludes the paper.

\begin{figure}[!t]
  \centering
  \includegraphics[scale=0.23]{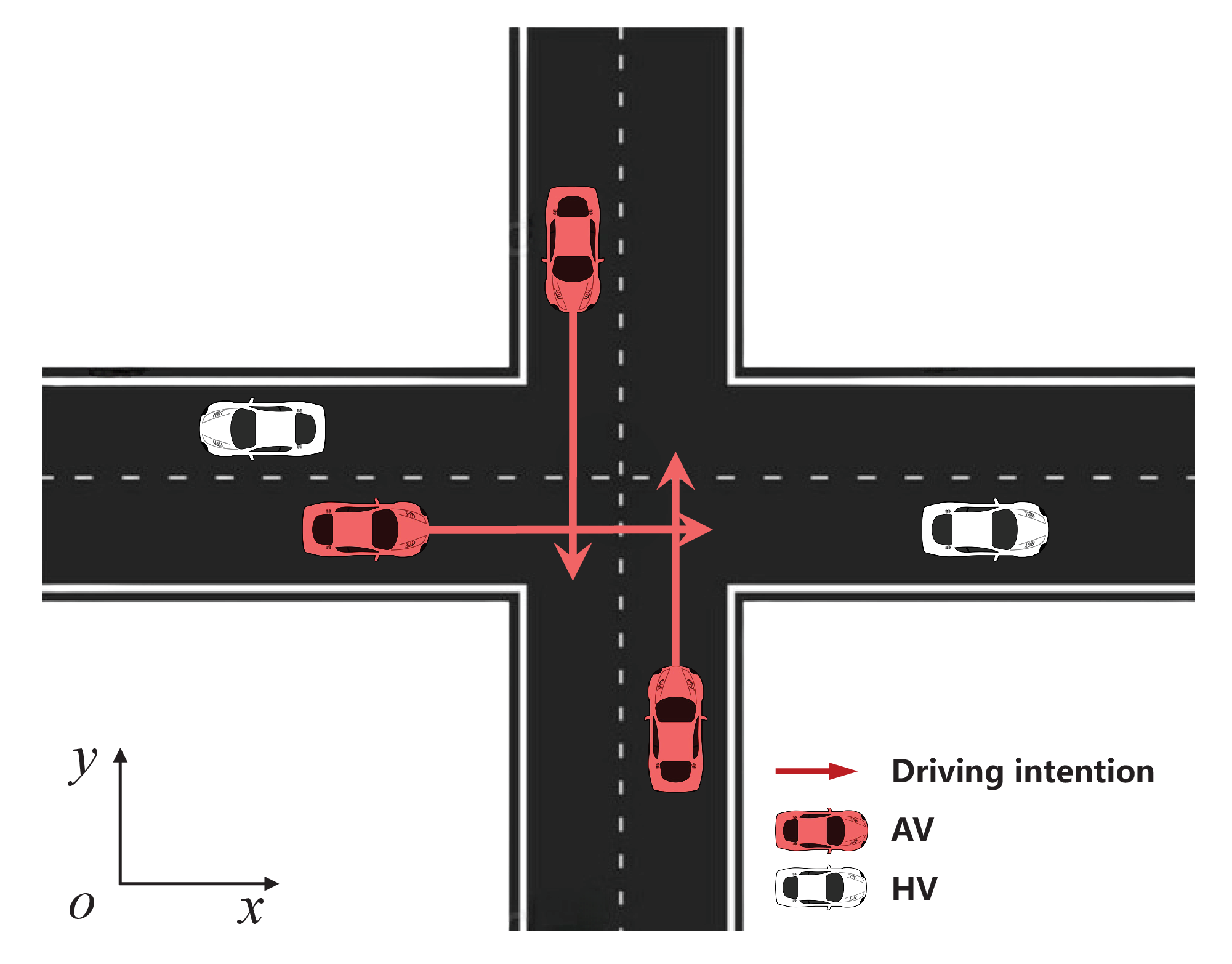}
  \caption{A multi-vehicle interaction scenario at an unsignalized intersection.}
  \label{fig.case}
\end{figure}

\section{Problem statement and system overview}
\label{Problem statement and system overview}

\begin{figure}[!tp]
  \centering
  \includegraphics[scale=0.20]{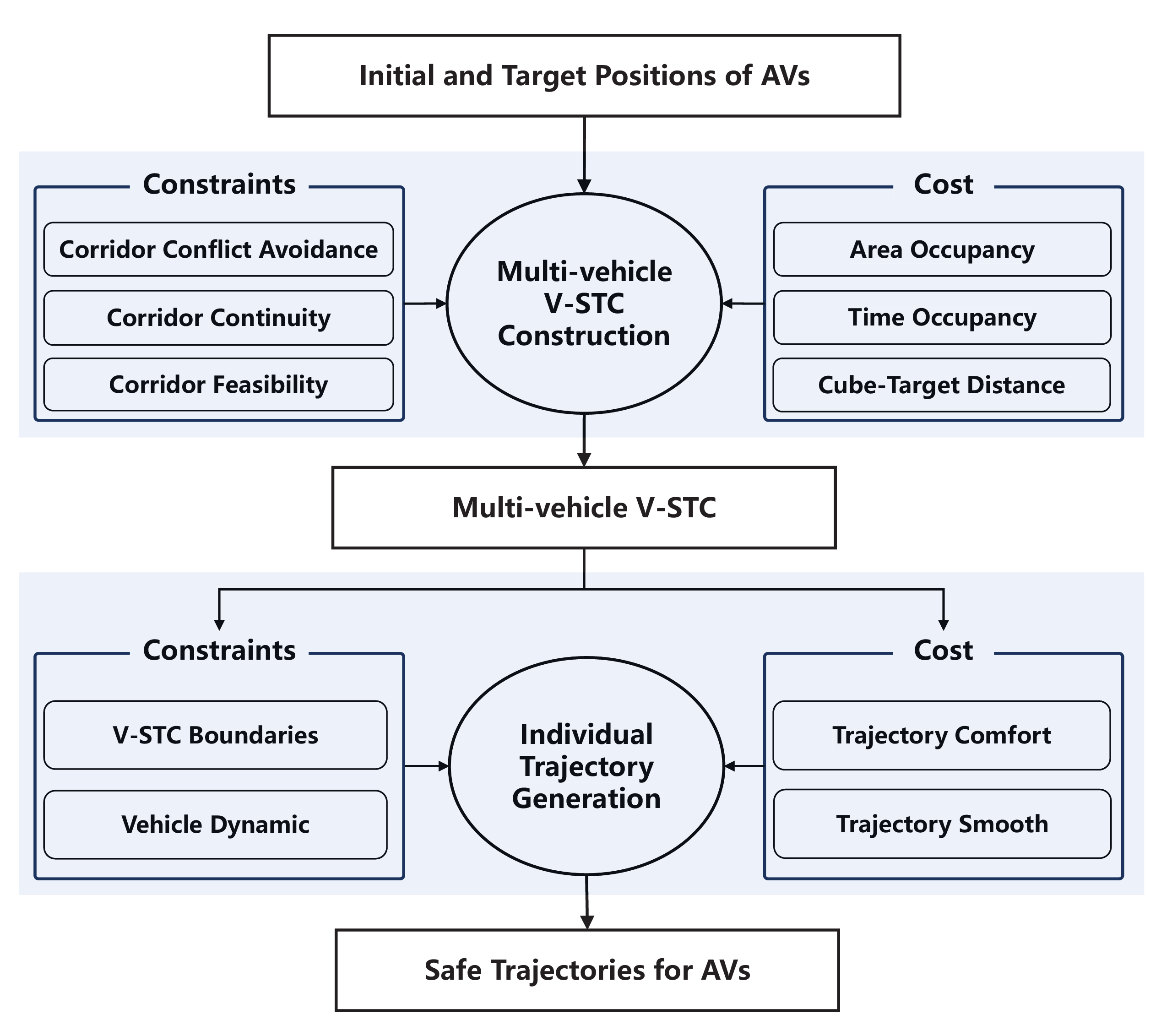}
  \caption{Overall framework of the proposed V-STC-based coordinated trajectory planning method.}
  \label{fig.framework}
\end{figure}

This work considers the coordinated trajectory planning problem for multiple AVs operating in shared environments where their intended routes may intersect, leading to potential collision risks. The objective is to compute a set of collision-free and dynamically feasible trajectories that drive all AVs from their initial positions to their assigned target positions while ensuring safety and efficiency. A representative interaction scenario at an unsignalized intersection is illustrated in Fig. \ref{fig.case}, which includes several AVs and human-driven vehicles (HVs).

Let the vehicle set be $\mathcal{V} = \{1, 2, \dots, N\}$, where $N$ denotes the number of AVs. For each AV $i \in \mathcal{V}$, the initial and target positions are denoted by $(x_{\mathrm{initial}}^i, y_{\mathrm{initial}}^i)$ and $(x_{\mathrm{target}}^i, y_{\mathrm{target}}^i)$ respectively. Each AV is governed by its dynamic model and must satisfy associated motion constraints, including collision-avoidance constraints with other vehicles and static obstacles. The coordinated trajectory planning task is therefore to determine trajectories that guide all AVs to their target positions while remaining safe and dynamically feasible throughout the motion.

\begin{figure}[!bp]
  \centering
  \includegraphics[scale=0.14]{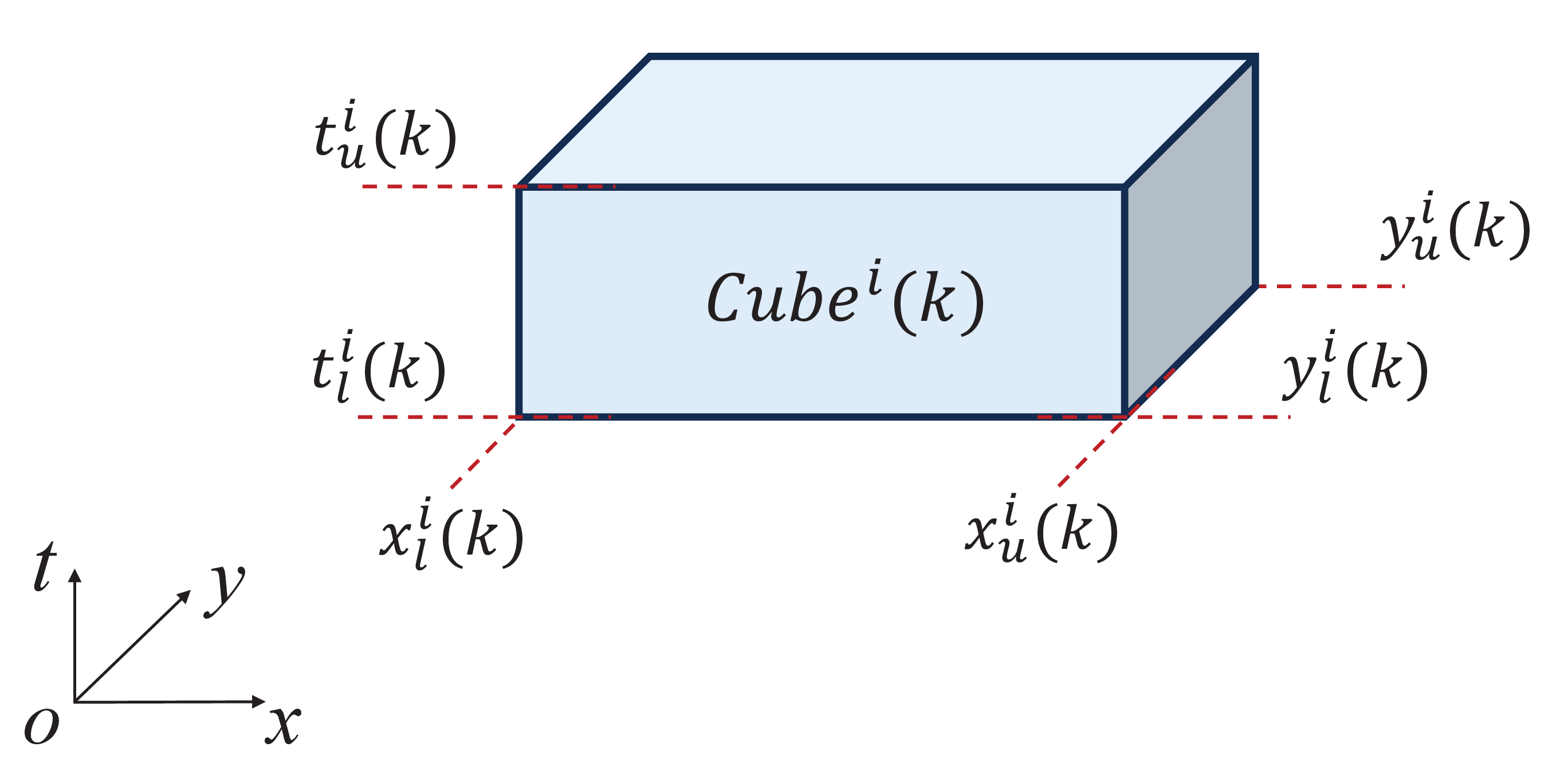}
  \caption{Spatio-temporal corridor cube $Cube^i(k)$ with spatial and temporal boundaries.}
  \label{fig.cube}
\end{figure}

To address this problem, we adopt a two-stage planning framework illustrated in Fig. \ref{fig.framework}. In the first stage, a V-STC is constructed for each AV, which consists of a sequence of linked spatio-temporal corridor cubes that form a collision-free admissible region in the joint $(x,y,t)$ domain. In the second stage, an individual trajectory is optimized within the corresponding V-STC by enforcing vehicle dynamics, boundary constraints, and trajectory-quality criteria.

\begin{figure*}[!t]
  \centering
  \subfloat[]{
  \includegraphics[width=3.3in]{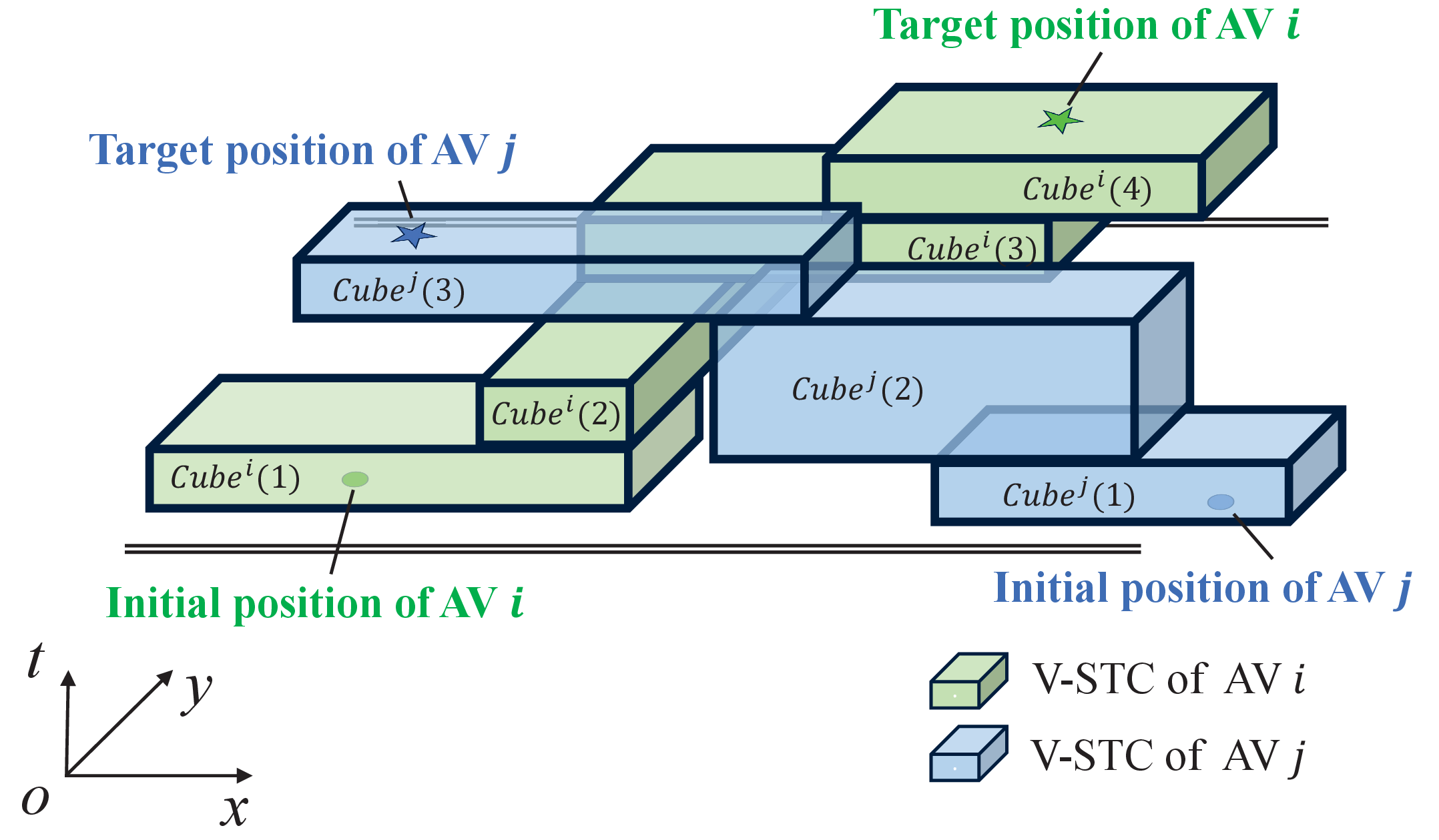}
  \label{fig.V-STC}}
  \quad
  \subfloat[]{
  \includegraphics[width=3.1in]{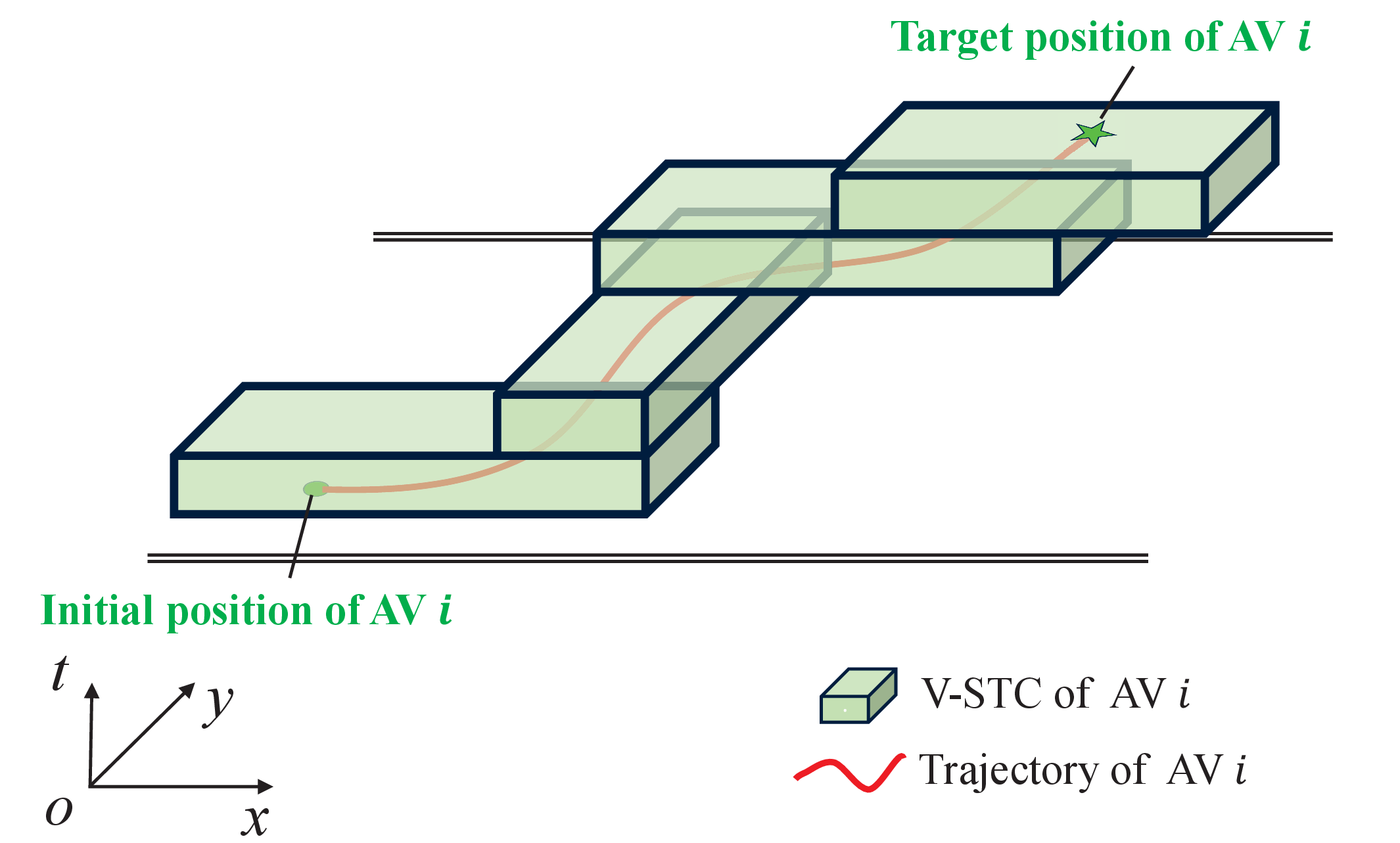}
  \label{fig.V-STC_trajectory}
  }
  \caption{Illustration of the V-STC and its use in the trajectory-planning framework. (a) Non-overlapping V-STC of two AVs. (b) The trajectory of AV $i$ planned inside its V-STC.}
  \label{fig.V-STC_case}
\end{figure*}

\section{Methodology}
\label{Methodology}
This section details the two-stage V-STC-based coordinated planning framework. First, we introduce the V-STC representation together with its geometric structure. Then, we formulate an optimization model to construct the V-STC for each AV, followed by a constrained trajectory optimization problem to generate smooth and dynamically feasible trajectories.

\subsection{Overview of the V-STC Representation}
The V-STC assigned to each AV is composed of a sequence of spatio-temporal corridor cubes. Each cube confines the admissible spatial region of the AV during a corresponding time interval. The geometric structure of a single corridor cube is shown in Fig. \ref{fig.cube}, which depicts its lower and upper spatial boundaries $\left[x^i_l(k), x^i_u(k) \right]$, $\left[y^i_l(k), y^i_u(k)\ \right]$ and temporal boundaries $\left[t^i_l(k), t^i_u(k) \right]$. For AV $i$, the $k$-th cube is defined as $Cube^i(k) = \left[x^i_l(k), x^i_u(k), y^i_l(k), y^i_u(k), t^i_l(k), t^i_u(k)\right], i \in \mathcal{V}, k \in \mathcal{K}$, where $\mathcal{K}=\{1, 2, \dots, K\}$ denotes the index set of corridor cubes.  

An AV's V-STC is constructed by linking a sequence of such cubes so that they form a continuous and feasible channel toward its target position. Fig. \ref{fig.V-STC_case}\subref{fig.V-STC} illustrates the V-STC of two AVs, showing their non-overlapping spatio-temporal layouts. These layouts encode the corridor-level collision-avoidance relationships between AVs in the interaction region. Fig. \ref{fig.V-STC_case}\subref{fig.V-STC_trajectory} provides a conceptual illustration of a trajectory of AV $i$ planned inside its V-STC, corresponding to the second stage of the framework after the corridor has been constructed.

\subsection{V-STC Construction}
The V-STC construction for each AV is formulated as a mixed-integer optimization problem. The objective is to determine the spatial layout and time durations of the corridor cubes while satisfying safety, continuity, feasibility, and boundary constraints. The optimization includes the following components.

\paragraph{Corridor Conflict Avoidance} 

\begin{figure}[!bp]
  \centering
  \includegraphics[scale=0.23]{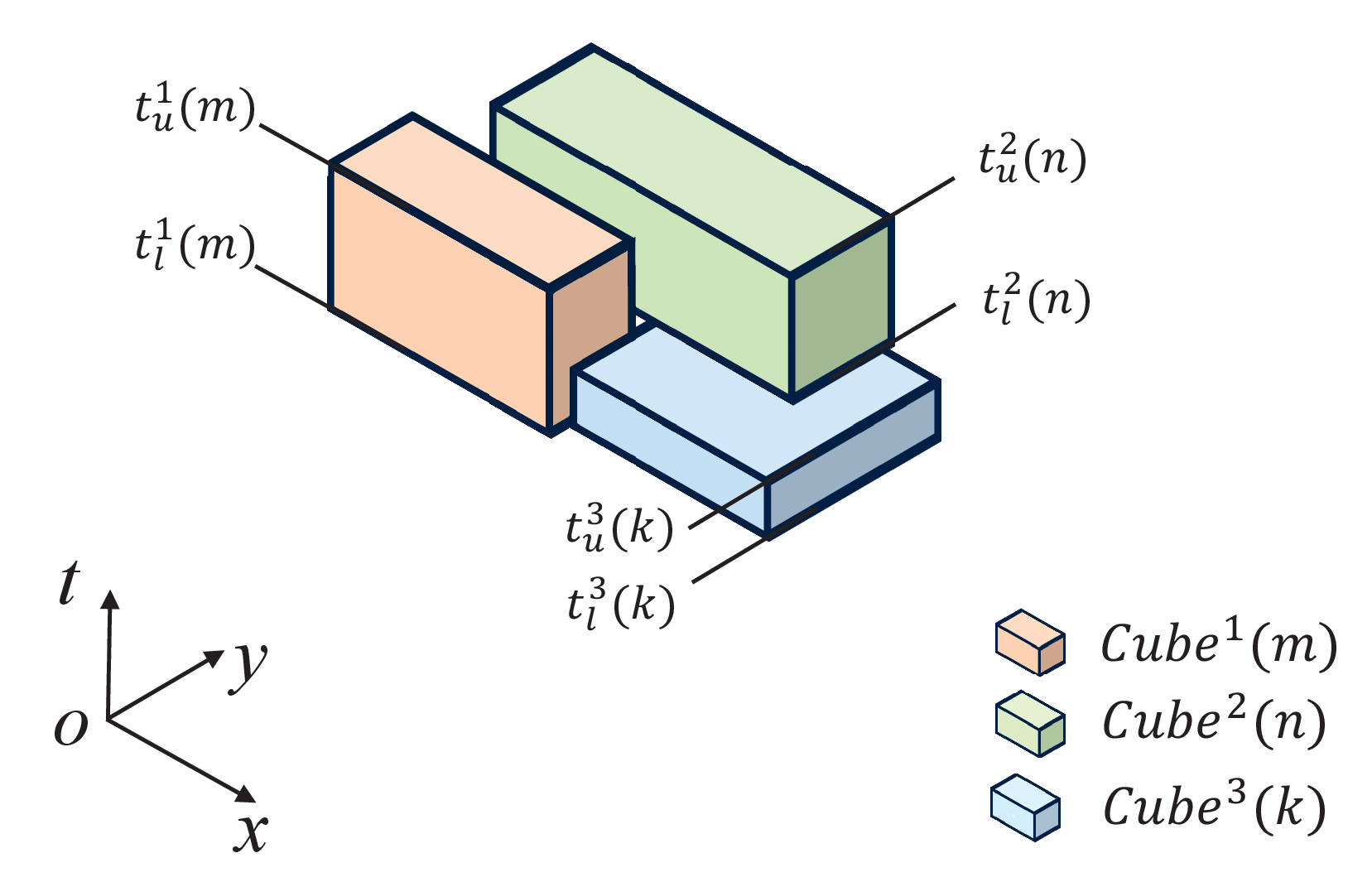}
  \caption{Illustration of temporal overlap among corridor cubes of different AVs, for which spatial separation must be enforced.}
  \label{fig.avoid_time}
\end{figure} 

To guarantee safety, any two corridor cubes belonging to different AVs must not overlap in the joint spatial and temporal domain. To encode the required logical relation that either the cubes are separated in time or they must be separated in space, three binary decision variables are introduced for each pair of cubes:
\begin{equation*}
  \begin{split}
    \delta^{ij}_x(m,n) &= 
    \begin{cases} 
      1, &  if \ x^i_u(m) \leq x^j_l(n), \\
      0, &  otherwise,
    \end{cases} \\
    \delta^{ij}_y(m,n) &= 
    \begin{cases} 
      1, &  if \ y^i_u(m) \leq y^j_l(n), \\
      0, &  otherwise,
    \end{cases} \\
    \delta^{ij}_t(m,n) &= 
    \begin{cases} 
      1, &  if \ \big[ t^i_l(m), t^i_u(m) \big] \cap \big[t^j_l(n), t^j_u(n)\big]=\emptyset,   \\
      0, &  otherwise,
    \end{cases} 
  \end{split}
\end{equation*}
where $\delta^{ij}_x(m,n)$ indicates whether spatial separation between $Cube^i(m)$ and $Cube^j(n)$ along the $x$-direction is active, $\delta^{ij}_y(m,n)$ indicates whether spatial separation between $Cube^i(m)$ and $Cube^j(n)$ along the $y$-direction is active, and $\delta^{ij}_t(m,n)$ indicates whether temporal separation between $Cube^i(m)$ and $Cube^j(n)$ is active. 

The collision-avoidance constraints are written as:
\begin{equation}
    \begin{split}
     &x^i_{l}(m) - x^j_{u}(n) + \delta^{ij}_x(m,n)M + \delta^{ij}_t(m,n)M \geq \gamma_x, \\
     &x^j_{l}(n) - x^i_{u}(m) + \delta^{ji}_x(n,m)M + \delta^{ij}_t(m,n)M \geq \gamma_x, \\
     &y^i_{l}(m) - y^j_{u}(n) + \delta^{ij}_y(m,n)M + \delta^{ij}_t(m,n)M \geq \gamma_y, \\
     &y^j_{l}(n) - y^i_{u}(m) + \delta^{ji}_y(n,m)M + \delta^{ij}_t(m,n)M \geq \gamma_y, \\
     &\delta^{ij}_x(m,n) + \delta^{ji}_x(n,m) + \delta^{ij}_y(m,n) + \delta^{ji}_y(n,m)  \\
     &\leq 3 + \delta^{ij}_t(m,n)M, \ \ \ \forall i,j \in \mathcal{V}, \forall m,n \in \mathcal{K}, \\
    \end{split}
    \label{avoi_AV}
\end{equation}
where $\gamma_x$ and $\gamma_y$ are the safe distances along the $x$-axis and $y$-axis, respectively, and $M$ is a sufficiently large positive number. 

\begin{figure}[!tp]
  \centering
  \includegraphics[scale=0.2]{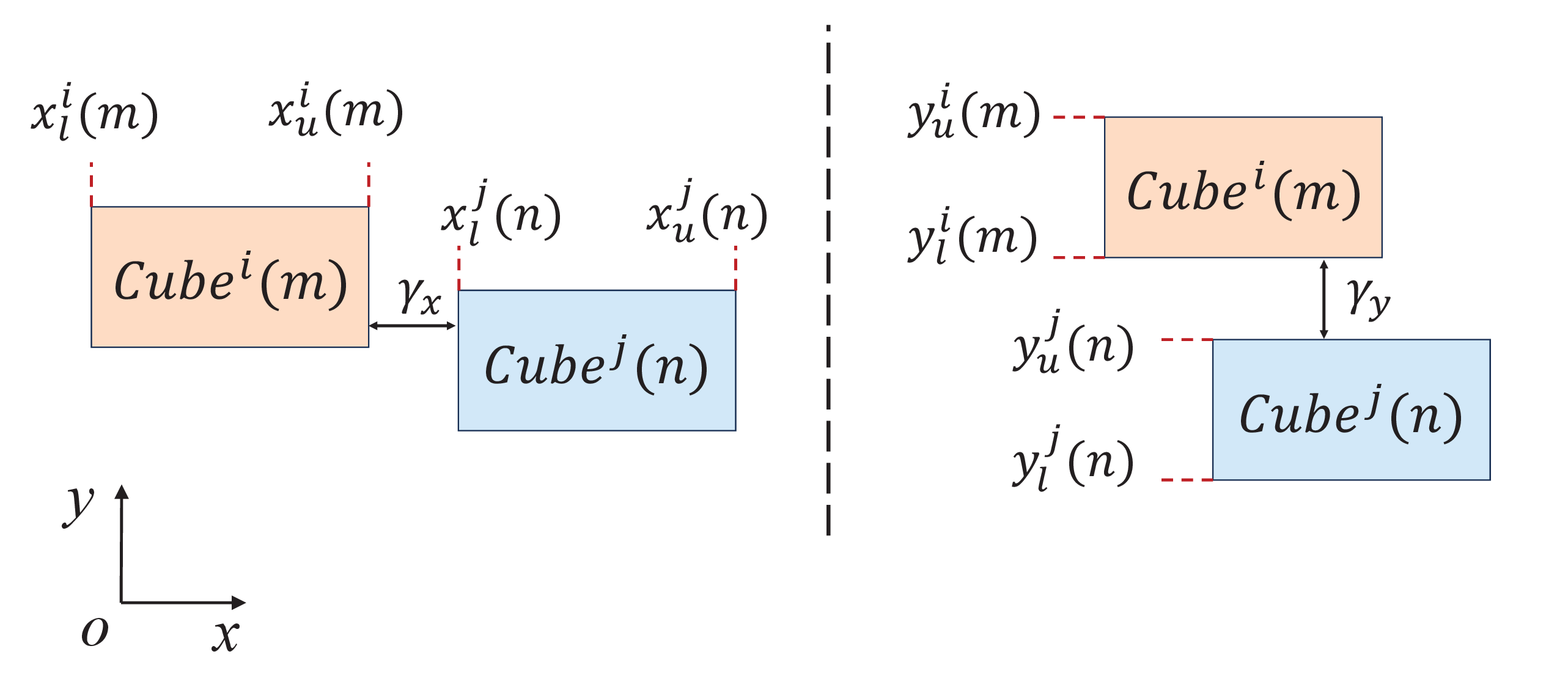}
  \caption{Required spatial separation between corridor cubes of different AVs that overlap temporally, illustrated along the $x$ and $y$ directions.}
  \label{fig.avoid_AV}
\end{figure}

These inequalities \eqref{avoi_AV} follow the standard Big-M structure \cite{Ding2014}. The binary variables activate or deactivate spatial and temporal separation terms. For any two corridor cubes belonging to different AVs that overlap in time, spatial separation must be enforced, as shown in Fig. \ref{fig.avoid_time} and \ref{fig.avoid_AV}.

\begin{figure}[htp]
  \centering
  \includegraphics[scale=0.25]{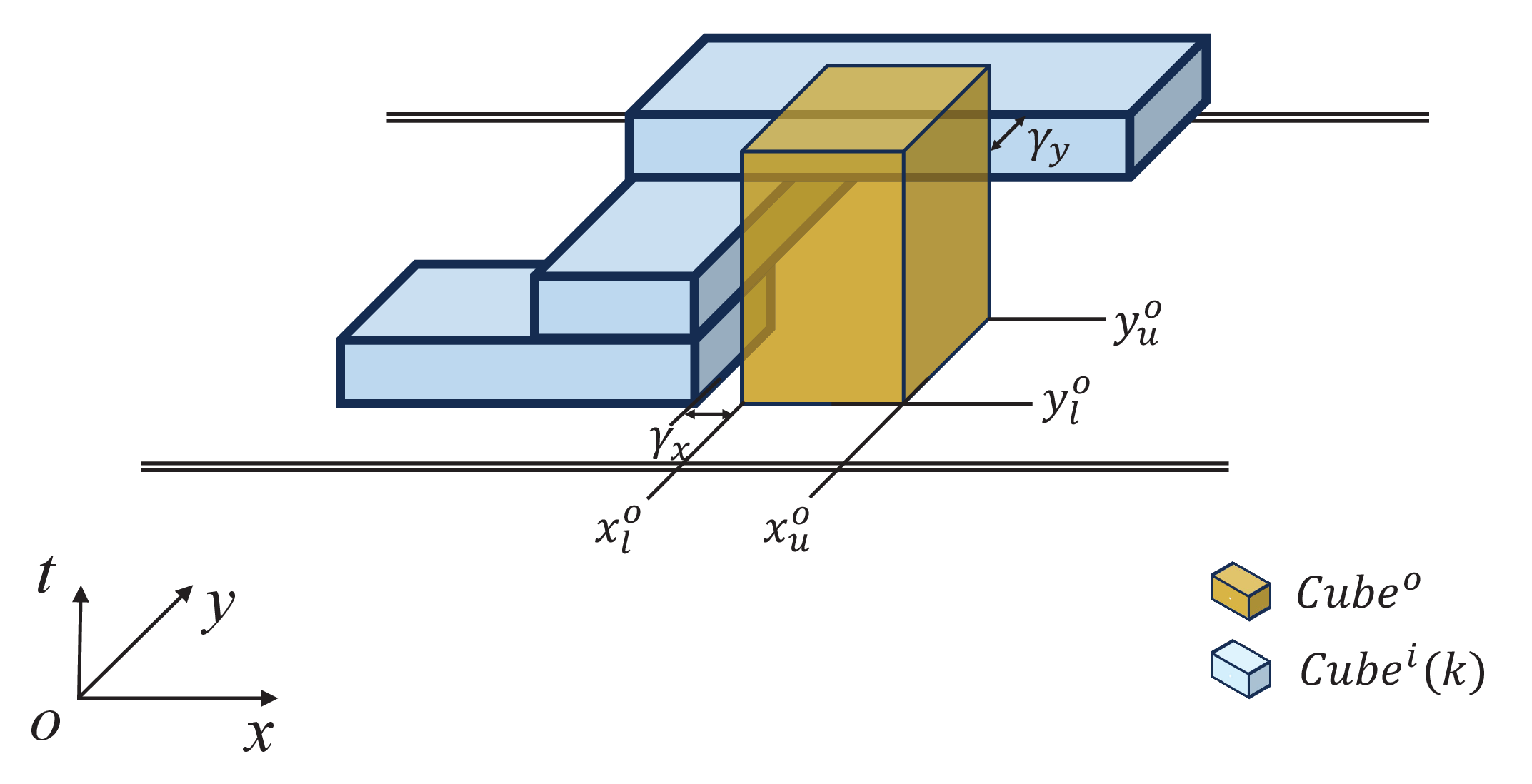}
  \caption{Geometric illustration of the collision-avoidance relationship between an AV's corridor cubes and static obstacles.}
  \label{fig.avoid_obs}
\end{figure}

Static obstacles are modeled as spatio-temporal cubes $Cube^o$, as shown in Fig. \ref{fig.avoid_obs}. Since obstacles are stationary, only spatial separation is required. The collision-avoidance constraint between an AV cube and an obstacle cube has the same Big-M structure as the spatial component of constraint \eqref{avoi_AV}: 
\begin{equation}
  \begin{split}
   &x^i_{l}(m) - x^o_{u}    + \delta^{io}_x(m)M  \geq \gamma_x, \\
   &x^o_{l}    - x^i_{u}(m) + \delta^{oi}_x(m)M  \geq \gamma_x, \\
   &y^i_{l}(m) - y^o_{u}    + \delta^{io}_y(m)M  \geq \gamma_y, \\
   &y^o_{l}    - y^i_{u}(m) + \delta^{oi}_y(m)M  \geq \gamma_y, \\
   &\delta^{io}_x(m)\!+\!\delta^{oi}_x(m)\!+\!\delta^{io}_y(m)\!+\!\delta^{oi}_y(m)\! \leq \!3, \forall i \! \in \! \mathcal{V}, \forall m \!\in \!\mathcal{K},\\
  \end{split}
  \label{avoi_obs}
\end{equation}
where $x^o_{l}$ and $x^o_{u}$ represent the lower and upper boundary coordinates of $Cube^o$ in the $x$ direction, respectively, and $y^o_{l}$ and $y^o_{u}$ represent the lower and upper boundary coordinates of $Cube^o$ in the $y$ direction, respectively. The binary variables $\delta^{io}_x(m)$ and $\delta^{io}_y(m)$ represent the relative positional relationships between $Cube^i(m)$ and $Cube^o$, and their definitions are similar to those in \eqref{avoi_AV}. Note that road boundaries are modeled as static obstacles to ensure AV safety.

\paragraph{Corridor Continuity}
To maintain the kinematic feasibility of the constructed spatio-temporal corridor, connectivity constraints are imposed between adjacent cubes. Let $\Delta t = t^i_u(k) - t^i_l(k)$ denote the time duration of $Cube^i(k)$, and let $x^i_c(k) = \frac{x^i_l(k) + x^i_u(k)}{2}$ and $y^i_c(k) = \frac{y^i_l(k) + y^i_u(k)}{2}$ denote the coordinates of the centroid of $Cube^i(k)$. The connectivity constraints are formulated by requiring that the reachable set of the current cube center over $\Delta t$ be contained within the subsequent cube:
\begin{equation}
    \begin{split}
      & x^i_{l}{(k \! + \! 1)} \! \leq \! x^i_c(k) \! + \! \Delta t \! \left[v_{\mathrm{min}}\delta^i_x(k) \! + \! v_{\mathrm{max}}(\delta^i_x(k) \! - \! 1)\right], \\
      & x^i_{u}{(k \! + \! 1)} \! \geq \! x^i_c(k) \! + \! \Delta t \! \left[v_{\mathrm{max}}\delta^i_x(k) \! + \! v_{\mathrm{min}}(\delta^i_x(k) \! - \! 1)\right], \\
      & y^i_{l}{(k \! + \! 1)} \! \leq \! y^i_c(k) \! + \! \Delta t \! \left[v_{\mathrm{min}}\delta^i_y(k) \! + \! v_{\mathrm{max}}(\delta^i_y(k) \! - \! 1)\right], \\
      & y^i_{u}{(k \! + \! 1)} \! \geq \! y^i_c(k) \! + \! \Delta t \! \left[v_{\mathrm{max}}\delta^i_y(k) \! + \! v_{\mathrm{min}}(\delta^i_y(k) \! - \! 1)\right], \\
      & \forall i \in \mathcal{V}, \forall k \in \mathcal{K}, \\
    \end{split}
    \label{contin_AV}
\end{equation}
where $v_{\mathrm{max}}$ and $v_{\mathrm{min}}$ are the maximum and minimum cruising velocities of the AVs, respectively, and $\delta^i_x(k)$, $\delta^i_y(k)$ are binary variables that indicate the relative direction of motion when transitioning from $Cube^i(k)$ to $Cube^i(k+1)$, defined as:
\begin{equation*}
  \begin{split}
    \delta^i_x(k) &= 
    \begin{cases} 
      1, &  if \ x^i_c(k) \leq x^i_c(k+1), \\
      0, &  otherwise,
    \end{cases} \\
    \delta^i_y(k) &= 
    \begin{cases} 
      1, &  if \ y^i_c(k) \leq y^i_c(k+1), \\
      0, &  otherwise.
    \end{cases} \\
  \end{split}
\end{equation*}

\begin{figure}[htp]
  \centering
  \includegraphics[scale=0.22]{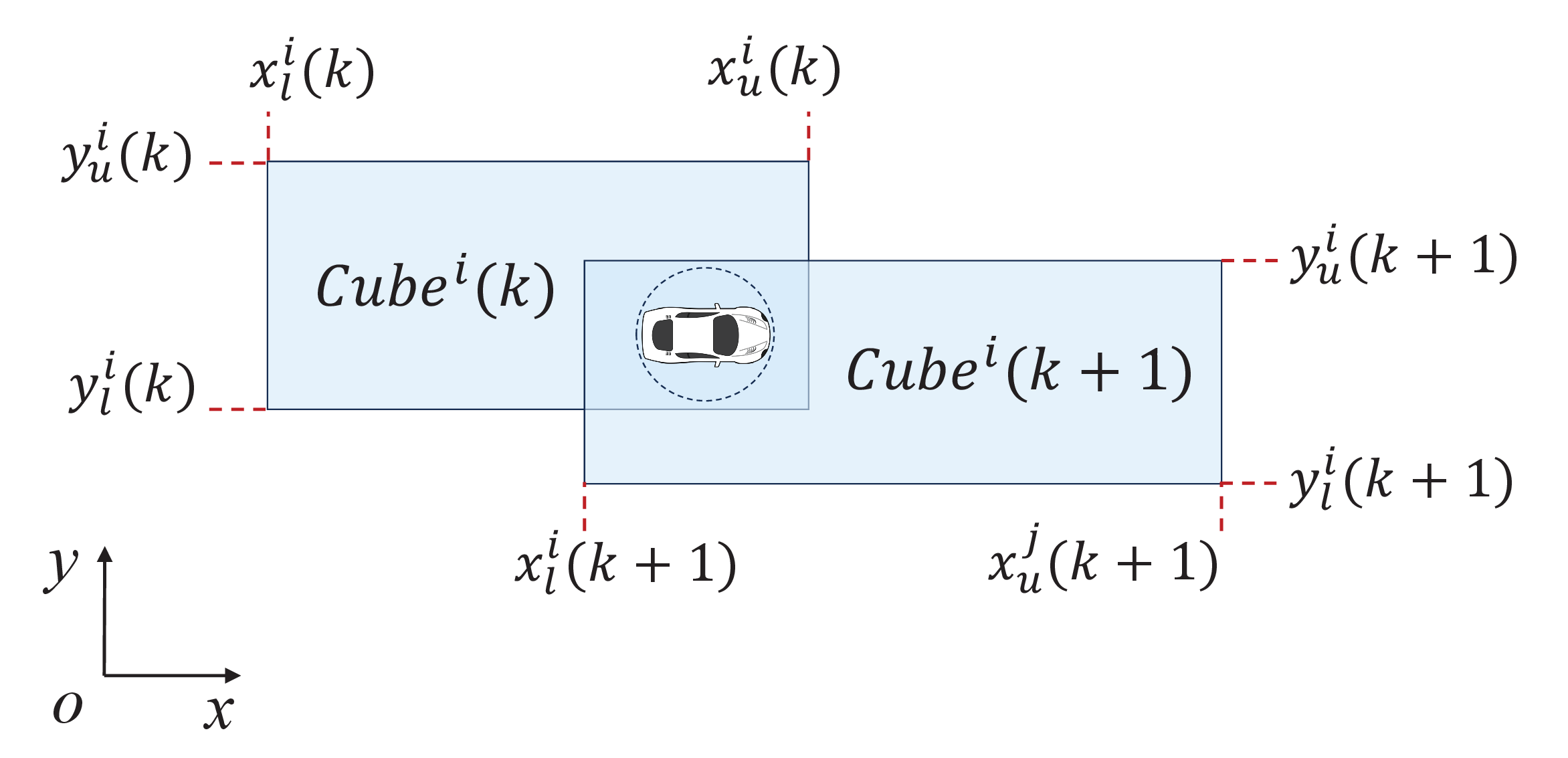}
  \caption{Overlapping area required between adjacent corridor cubes to ensure continuity.}
  \label{fig.overlap}
\end{figure}

To further guarantee feasible transitions between successive cubes, a minimum spatial overlap must exist between $Cube^i(k)$ and $Cube^i(k+1)$, as shown in Fig. \ref{fig.overlap}. This overlap prevents discontinuities and ensures that the AV can move smoothly into the next cube. The overlap constraints are formulated as:
\begin{equation}
    \begin{split}
     & x^i_{u}(k) - x^i_{l}(k+1) \geq \gamma_r, \\
     & x^i_{u}(k+1) - x^i_{l}(k) \geq \gamma_r, \\
     & y^i_{u}(k) - y^i_{l}(k+1) \geq \gamma_r, \\
     & y^i_{u}(k+1) - y^i_{l}(k) \geq \gamma_r, \ \ \forall i \in \mathcal{V}, \forall k \in \mathcal{K},\\
    \end{split}
    \label{cube_overlap}
\end{equation}
where $\gamma_r = \sqrt{l^2 + w^2}$ is determined by the vehicle length $l$ and width $w$.

\paragraph{Corridor Feasibility}
Each corridor cube is required to have sufficient spatial extent to contain the AV footprint, and its duration is bounded within a prescribed interval. These are formulated as follows:
\begin{equation}
    \begin{split}
    & x^i_u(k) - x^i_l(k) \geq \gamma_r, \\
    & y^i_u(k) - y^i_l(k) \geq \gamma_r, \\
    & t_{\mathrm{min}} \leq t^i_u(k) - t^i_l(k) \leq t_{\mathrm{max}}, \ \ \forall i \in \mathcal{V}, \forall k \in \mathcal{K},\\
  \end{split}
  \label{accomm_cube}
\end{equation}
where $t_{\mathrm{min}}$ and $t_{\mathrm{max}}$ represent the minimum and maximum time steps of each corridor cube, respectively.

Furthermore, to guarantee that the constructed spatio-temporal corridor connects the initial and target positions, the first corridor cube is required to contain the AV's initial position, and the last corridor cube is required to contain its target position. This requirement is enforced by:
\begin{equation}
  \begin{split}
      & x^i_{l}(1) + \frac{\gamma_r}{2} \leq  x^i_{\mathrm{initial}}  \leq x^i_{u}(1) - \frac{\gamma_r}{2}, \\
      & y^i_{l}(1) + \frac{\gamma_r}{2} \leq  y^i_{\mathrm{initial}}  \leq y^i_{u}(1) - \frac{\gamma_r}{2}, \\
      & x^i_{l}(K) + \frac{\gamma_r}{2} \leq  x^i_{\mathrm{target}}  \leq x^i_{u}(K) - \frac{\gamma_r}{2}, \\
      & y^i_{l}(K) + \frac{\gamma_r}{2} \leq  y^i_{\mathrm{target}}  \leq y^i_{u}(K) - \frac{\gamma_r}{2}. \\
    \end{split}
    \label{int_cube_include}
\end{equation}

\paragraph{Optimization Model for V-STC Generation}
The V-STC construction is formulated as a mixed-integer optimization problem that determines both the spatial layout and the temporal duration of the corridor cubes. Specifically, we define three cost terms for each AV $i$ and cube index $k$, i.e., a goal-reaching term that penalizes deviation of the cube centroid from the target position, a spatial-size term that encourages larger corridor cubes, and a temporal term that penalizes long corridor durations, given by: 

\begin{equation}
  \label{cost}
  \begin{split}
    & J^i_{\mathrm{goal}}(k) =  \left(x^i_c(k) - x_{\mathrm{target}}^i\right)^2 + \left(y^i_c(k) - y_{\mathrm{target}}^i\right)^2, \\
    & J^i_{\mathrm{size}}(k) = x^i_u(k) - x^i_l(k) + y^i_u(k) - y^i_l(k), \\
    & J^i_{\mathrm{time}}(k) = t^i_u(k) - t^i_l(k). \\
  \end{split}
\end{equation}

The overall objective combines the above cost terms over all AVs and cubes, and the resulting optimization problem is formulated as:
\begin{equation}
  \begin{split}
  \label{MIQP}
  & \underset {C^1, \dots, C^N} {min} \sum_{i=1}^{N} \sum_{k=1}^{K} \left[\omega_g J^i_{\mathrm{goal}}(k) - \omega_s J^i_{\mathrm{size}}(k) + \omega_t J^i_{\mathrm{time}}(k) \right] \\
  & \ \ \ subject \ to \ (1)\thicksim (6),\\
\end{split}
\end{equation}
where $C^i = \left[ Cube^i(1), Cube^i(2), \dots, Cube^i(K) \right]$, and $\omega_g$, $\omega_s$, $\omega_t$ are weighting coefficients.

The designed optimization model \eqref{MIQP} can be solved via the branch-and-bound method \cite{Nataraj2011}, generating a V-STC containing $K$ corridor cubes for each AV.

\paragraph{Minimum-Cube V-STC Generation}

To reduce the overall temporal occupancy of the V-STC, it is desirable to construct the corridor using only those cubes that are necessary for connecting the AV's initial and target positions. Due to the cost term $J^i_{\mathrm{goal}}$ in \eqref{MIQP}, which encourages each cube to be placed near the target, the solution of \eqref{MIQP} may include more cubes than are actually required for forming a valid corridor. In particular, cubes that appear after the target has already been reached no longer contribute to the corridor structure and should be removed.

\begin{figure}[htp]
  \centering
  \includegraphics[scale=0.2]{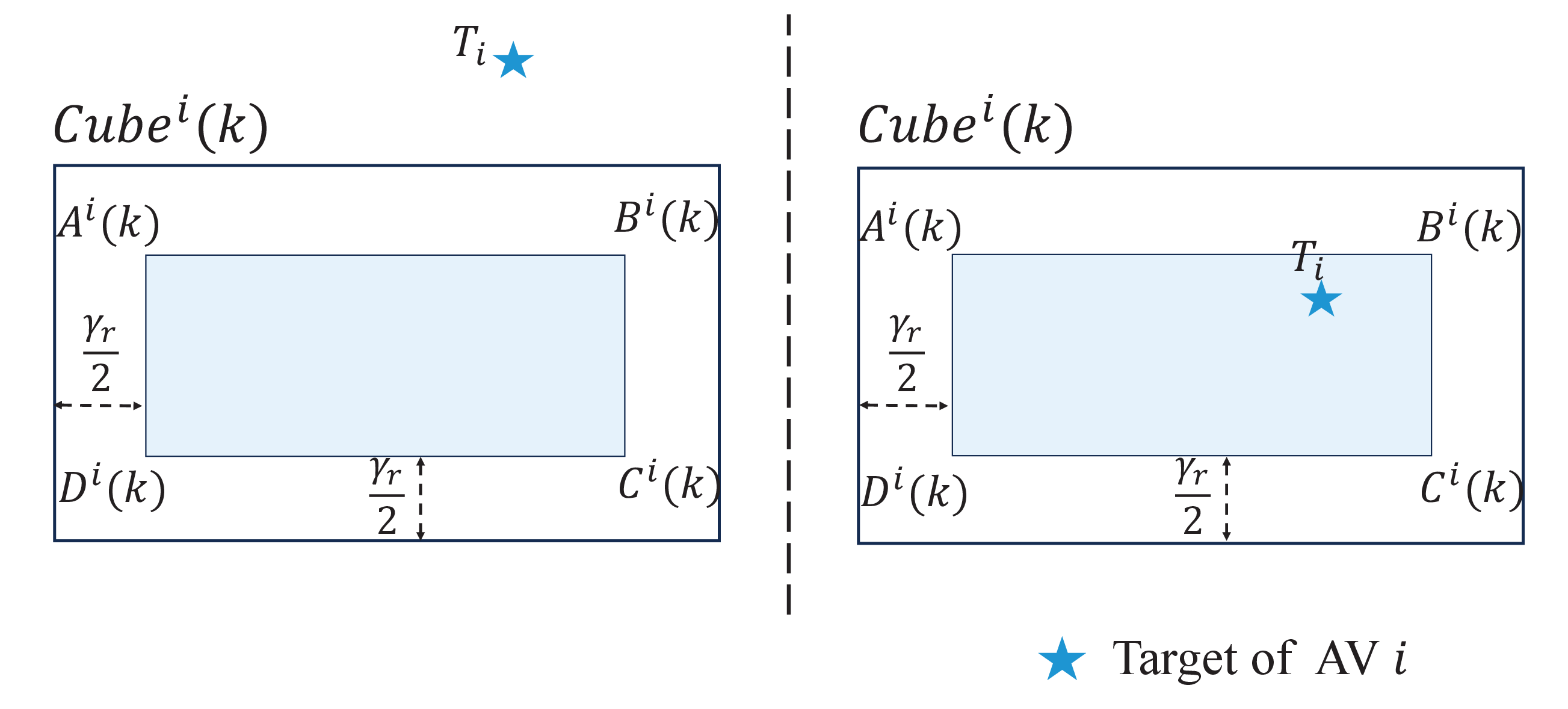}
  \caption{Relative positioning of corridor cubes and target position.}
  \label{fig.num_cube}
\end{figure}

To extract a shortened V-STC with the minimum number of cubes from the optimized solution, we examine the corridor cubes sequentially from the initial position toward the goal. For each cube $Cube^i(k)$, an inward-shrunk axis-aligned rectangle $R^i(k)$, as shown in Fig. \ref{fig.num_cube}, is constructed as:
\begin{equation*}
  R^i(k) = conv\{A^i(k), B^i(k), C^i(k), D^i(k)\},
\end{equation*}
where the four vertices are given by
\begin{equation*}
  \begin{cases} 
    \begin{split}
      &A^i_{x}(k) = x_l^i(k) + \frac{\gamma_r}{2}, \ \ A^i_{y}(k) = y_u^i(k) - \frac{\gamma_r}{2}, \\
      &B^i_{x}(k) = x_u^i(k) - \frac{\gamma_r}{2}, \ \ B^i_{y}(k) = y_u^i(k) - \frac{\gamma_r}{2}, \\
      &C^i_{x}(k) = x_u^i(k) - \frac{\gamma_r}{2}, \ \ C^i_{y}(k) = y_l^i(k) + \frac{\gamma_r}{2}, \\
      &D^i_{x}(k) =  x_l^i(k) + \frac{\gamma_r}{2}, \ \ D^i_{y}(k) = y_l^i(k) + \frac{\gamma_r}{2}. \\
    \end{split} 
  \end{cases}
\end{equation*}

We then check whether the goal position $T^i$ is contained in $R^i(k)$. The cubes are tested in increasing order of $k$, and the first index $k^{\star}$ satisfying $T^i \in R^i(k^{\star})$ is selected as the terminal cube. All cubes with indices $k > k^{\star}$ are removed, and the remaining cubes $Cube^i(1), \dots ,Cube^i(K^i)$ form the pruned V-STC for AV $i$, with $K^i = k^{\star}$.

\begin{remark}
 \label{remark1}
 Compared with existing STC-based approaches \cite{Zhang2023, Zang2023, Zang2024}, the proposed method introduces two key improvements. First, it eliminates the reliance on predefined reference waypoints. In conventional STC methods, each waypoint corresponds to a corridor cube, thereby implicitly determining both the number and spatial arrangement of corridor cubes. In contrast, the proposed V-STC allows both the number and placement of cubes to be optimized directly, resulting in a more flexible spatial structure. Second, instead of enforcing fixed time steps for all corridor cubes as in \cite{Zhang2023, Zang2023, Zang2024}, the proposed method treats the duration of each cube as a decision variable. This adaptive temporal modeling reduces overall corridor occupancy time and improves multi-vehicle coordination efficiency, particularly in constrained or conflict-prone scenarios. Overall, conventional STC formulations are sensitive to the selection of reference waypoints and time-step parameters, with feasibility and efficiency typically relying on careful parameter tuning and inappropriate settings potentially leading to infeasible or inefficient solutions, whereas the proposed approach jointly optimizes both spatial and temporal structures, thereby eliminating the need for manual parameter adjustment and further improving coordination efficiency.
\end{remark}

\subsection{Trajectory Generation Within the V-STC}
Given the V-STC assigned to each AV, the coordinated multi-vehicle planning problem is decomposed into independent trajectory optimization problems, where each AV generates a dynamically feasible trajectory within its own corridor. For AV $i$, the corridor time horizon $[0, t^i_u(K^i)]$ is discretized into $T$ uniform intervals with step size $\delta t =\frac{t^i_u(K^i)}{T}$. A constrained optimal control problem is then solved to compute the trajectory.

\paragraph{Kinematic Constraints}
  The trajectory must satisfy the AV's kinematic constraints, which are formulated as:
  \begin{equation}
    \label{model}
    \begin{cases} 
      \begin{split}
      & x^i(t \! + \! 1) = x^i(t) \! + \! v^i(t)\cos \psi^i(t) \delta t \\
      & y^i(t \! + \! 1) = y^i(t) \! + \! v^i(t)\sin \psi^i(t) \delta t \\
      & \psi^i(t \! + \! 1) = \psi^i(t) \! + \! v^i(t) \frac{\tan \phi^i(t)}{L_w} \delta t \\
      & v^i(t \! + \! 1) = v^i(t) + a^i(t) \delta t \\
      & a^i(t \! + \! 1) = a^i(t) + {jerk}^i(t) \delta t \\
      \end{split} 
    \end{cases},
  \end{equation}
  where $ \left( x^i(t),y^i(t) \right)$ denotes the position of AV $i$ at time step $t$, $\psi^i(t)$ is the heading angle, $v^i(t)$ is the longitudinal velocity, $a^i(t)$ is the longitudinal acceleration, $\phi^i(t)$ is the front-wheel steering angle, ${jerk}^i(t)$ is the jerk input, and $L_w$ is the wheelbase.

  The acceleration and steering angle are bounded by:
  \begin{equation}
    \begin{split}
      &a_{\mathrm{min}} \leq a^i(t) \leq a_{\mathrm{max}},\\
      &\phi_{\mathrm{min}} \leq \phi^i(t) \leq \phi_{\mathrm{max}}, \ \ \ \forall i \in \mathcal{V}, \\
    \end{split}
  \end{equation}
  where $a_{\mathrm{max}}$ and $a_{\mathrm{min}}$ represent the upper and lower bounds of the vehicles' acceleration, respectively, and $\phi_{\mathrm{max}}$ and $\phi_{\mathrm{min}}$ are the maximum and minimum front-wheel steering angles of the vehicles, respectively.

    \begin{figure}[htp]
      \centering
      \includegraphics[scale=0.22]{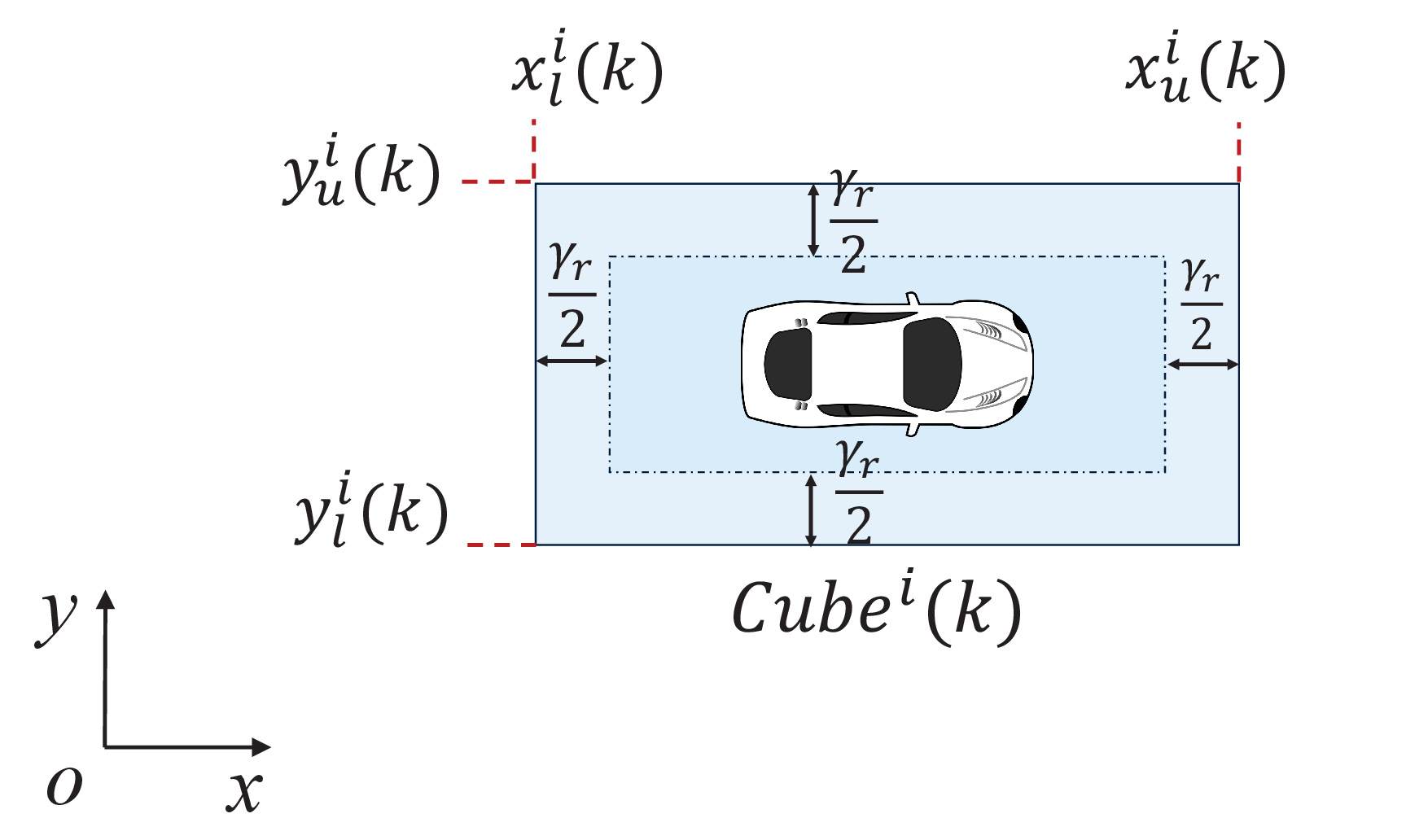}
      \caption{Boundary of the corridor cube.}
      \label{fig.in_cube}
    \end{figure}

  \paragraph{Boundary Conditions}
  Each AV starts from its initial position at an initial velocity $v^i_{\mathrm{initial}}$ and an orientation angle $\psi^i_{\mathrm{initial}}$, while both the initial acceleration and the front-wheel steering angle are set to zero. Accordingly, the boundary conditions for the initial and target states are given by:
  \begin{equation}
    \begin{split}
      &\left[x^i(0), y^i(0), \psi^i(0), v^i(0), a^i(0), \phi^i(0) \right] \\
      & = \left[x_{\mathrm{initial}}^i, y_{\mathrm{initial}}^i, \psi^i_{\mathrm{initial}}, v^i_{\mathrm{initial}}, 0, 0 \right],\\
      &\left[x^i(T), y^i(T) \right] = \left[ x_{\mathrm{target}}^i, y_{\mathrm{target}}^i\right].\\
    \end{split}
  \end{equation}

  \paragraph{Corridor Constraints}
  To ensure collision-free execution, the AV body is required to remain inside its assigned corridor at all time steps, as illustrated in Fig. \ref{fig.in_cube}. Let $ \mathcal{T}^i_k = \{ t | t \in \lfloor \frac{t^i_l(k)}{\delta t} , \frac{ t^i_u(k)}{\delta t} \rceil \}$. Then, for all $t\in \mathcal{T}^i_k$,
  \begin{equation}
    \begin{split}
      &x^i_{l}(k) + \frac{\gamma_r}{2} \leq x^i(t) \leq x^i_{u}(k) - \frac{\gamma_r}{2},\\
      &y^i_{l}(k) + \frac{\gamma_r}{2} \leq y^i(t) \leq y^i_{u}(k) - \frac{\gamma_r}{2}. \\
    \end{split}
  \end{equation}

\paragraph{Cost Function for Generating Trajectories}
We formulate the trajectory generation as a constrained optimal control problem with a composite objective consisting of comfort and smoothness terms. The comfort term penalizes acceleration and jerk:
\begin{equation*}
  J^i_{\mathrm{comfort}} = \sum_{t=0}^{T} \left[ ({jerk}^i(t))^2 + (a^i(t))^2 \right],
\end{equation*}
while the smoothness term penalizes changes in steering and heading angles:
\begin{equation*}
  J^i_{\mathrm{smooth}} = \sum_{t=1}^{T} \left[ (\psi^i(t) - \psi^i(t-1))^2 + (\phi^i(t) - \phi^i(t-1))^2 \right].
\end{equation*}

The resulting optimization problem is:
\begin{equation}
  \begin{split}
  \label{NLP}
  \underset {u^i(0), \dots, u^i(T)} {min} & \left( \omega_1 J^i_{\mathrm{comfort}} + \omega_2 J^i_{\mathrm{smooth}} \right) \\
  & subject \ to \ (10)\thicksim (13), \\
\end{split}
\end{equation}
where $u^i(t) = \left[ {jerk}^i(t), \phi^i(t) \right]$, and $\omega_1, \omega_2$ are tunable weights that balance comfort and smoothness.

The constrained optimal control problem in \eqref{NLP} is solved using a interior-point method (IPM) \cite{Pardo2016}. The optimal control sequence $\{u^i(t)\}^T_{t=0}$ is then applied to the discretized model in \eqref{model} to obtain the state trajectory $\{ x^i(t), y^i(t), \psi^i(t), v^i(t), a^i(t)\}^T_{t=0}$.

\begin{table}[htp]
  \caption{Relevant Parameters of The Simulation Experiments}
  \label{Parameter}
  \centering
    \begin{tabular}{|c|c|}
    \hline
    Parameters & Value  \\   \hline
    Safe distance $\gamma_x$, $\gamma_y$  & 0.1(m) \\   \hline
    Maximum cruising velocities $v_{\mathrm{max}}$      & 20(m/s) \\    \hline
    Minimum cruising velocities $v_{\mathrm{min}}$      & 0(m/s) \\   \hline
    Maximum front-wheel steering angle $\phi_{\mathrm{max}}$ & $\frac{\pi}{3}(rad)$ \\   \hline
    Minimum front-wheel steering angle $\phi_{\mathrm{min}}$ & $-\frac{\pi}{3}(rad)$ \\ \hline
    Maximum acceleration of the vehicle $a_{\mathrm{max}}$ & 4(m/$s^2$) \\   \hline
    Minimum acceleration of the vehicle $a_{\mathrm{min}}$ & 0(m/$s^2$) \\ \hline
    Maximum time step of corridor cube $t_{\mathrm{max}}$ & 1.0(s) \\   \hline
    Minimum time step of corridor cube $t_{\mathrm{min}}$ & 0.1(s) \\   \hline
    The length of the vehicle $l$ & $4(m)$ \\   \hline 
    The width of the vehicle $w$ & $2(m)$  \\   \hline 
    The wheelbase of the vehicle $L_w$ & $2.8(m)$ \\ \hline
  \end{tabular}
\end{table}

\section{Experimental Results and Discussion}
\label{Experimental Results and Discussion}
To evaluate the effectiveness of the proposed V-STC method, we compare it with the baseline STC method \cite{Zhang2023} across three representative scenarios: (a) unsignalized intersections, (b) multi-lane lane changes, and (c) unstructured environments. The parameter settings of V-STC are summarized in Table \ref{Parameter}, and the STC method uses the same settings unless otherwise specified. Specifically, STC requires additional settings for reference waypoints and the corridor time step. The waypoints are manually arranged along the driving direction with a nominal spacing of 10 $m$ (denoted by $\bigstar$), while the corridor time step is fixed at 1 $s$.

\begin{figure*}[hbp]
  \centering
  \subfloat[]{
  \includegraphics[width=3.3in]{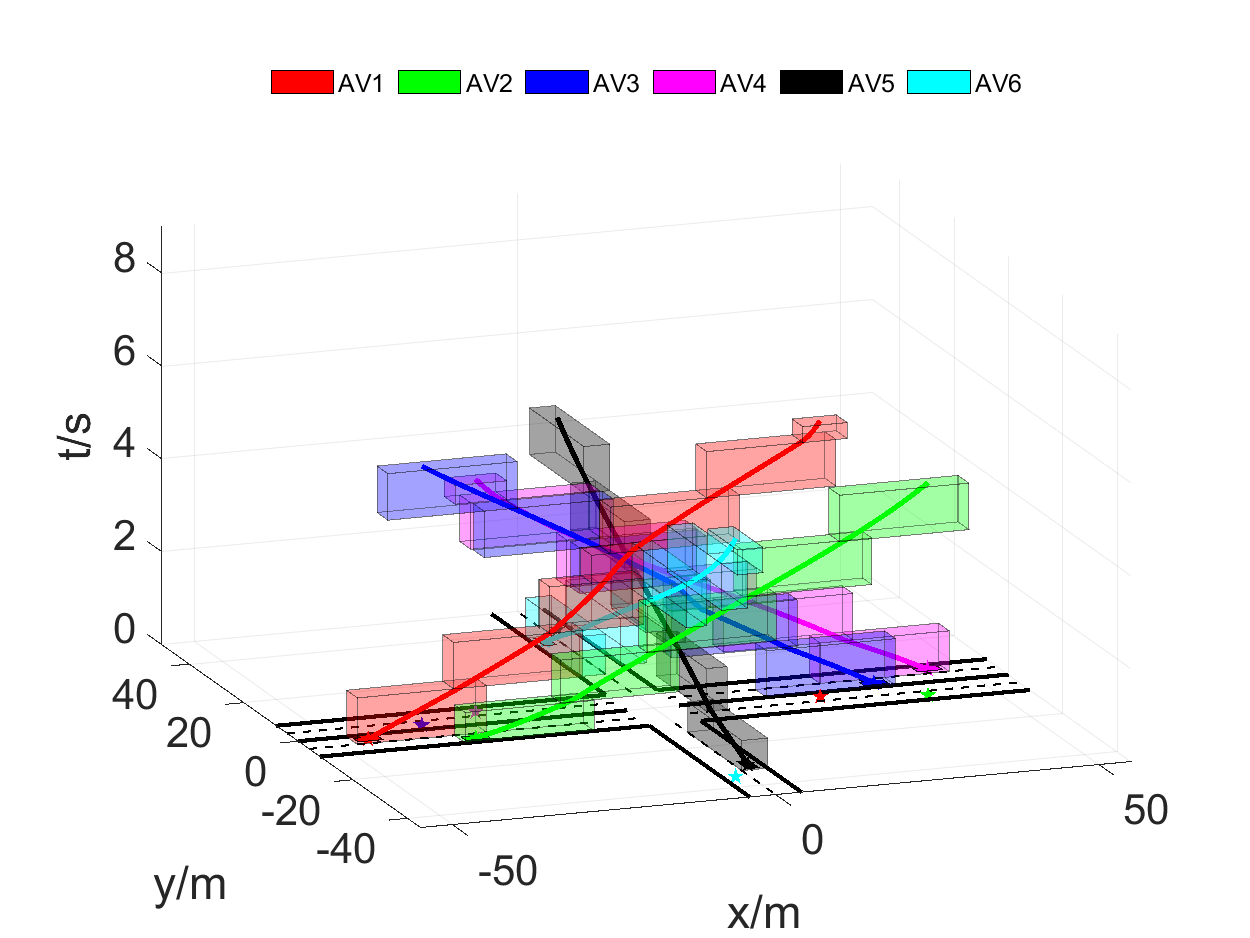}
  \label{fig.stc_auto_n1}}
  \quad
  \subfloat[]{
  \includegraphics[width=3.3in]{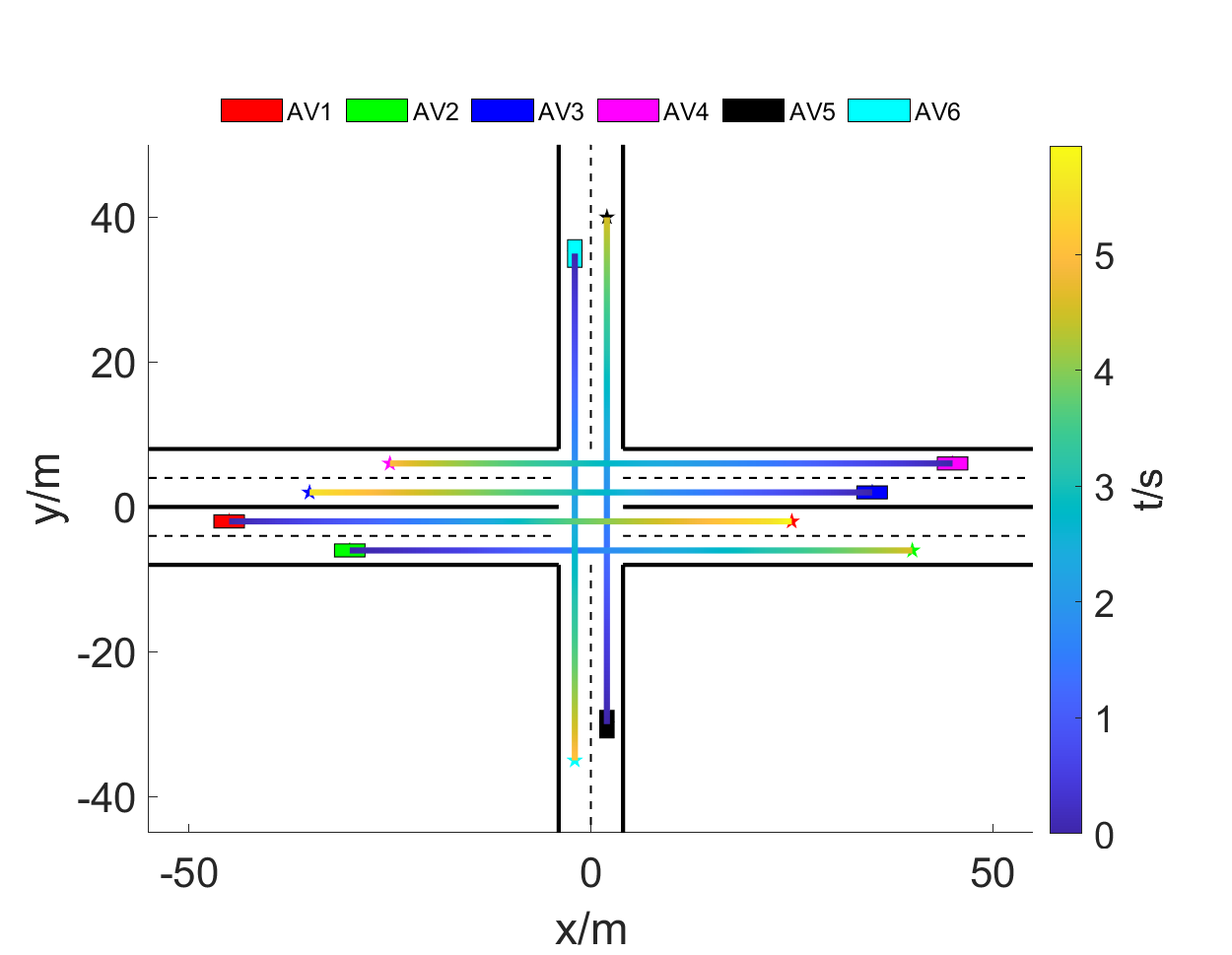}
  \label{fig.tra_auto_n1}
  }
  \caption{Simulation results of the proposed V-STC method in the unsignalized intersection scenario: (a) safety corridors; (b) trajectories.}
  \label{fig.stc_tra_auto_n1}
\end{figure*}

\begin{figure*}[hbp]
  \centering
  \subfloat[]{
  \includegraphics[width=3.3in]{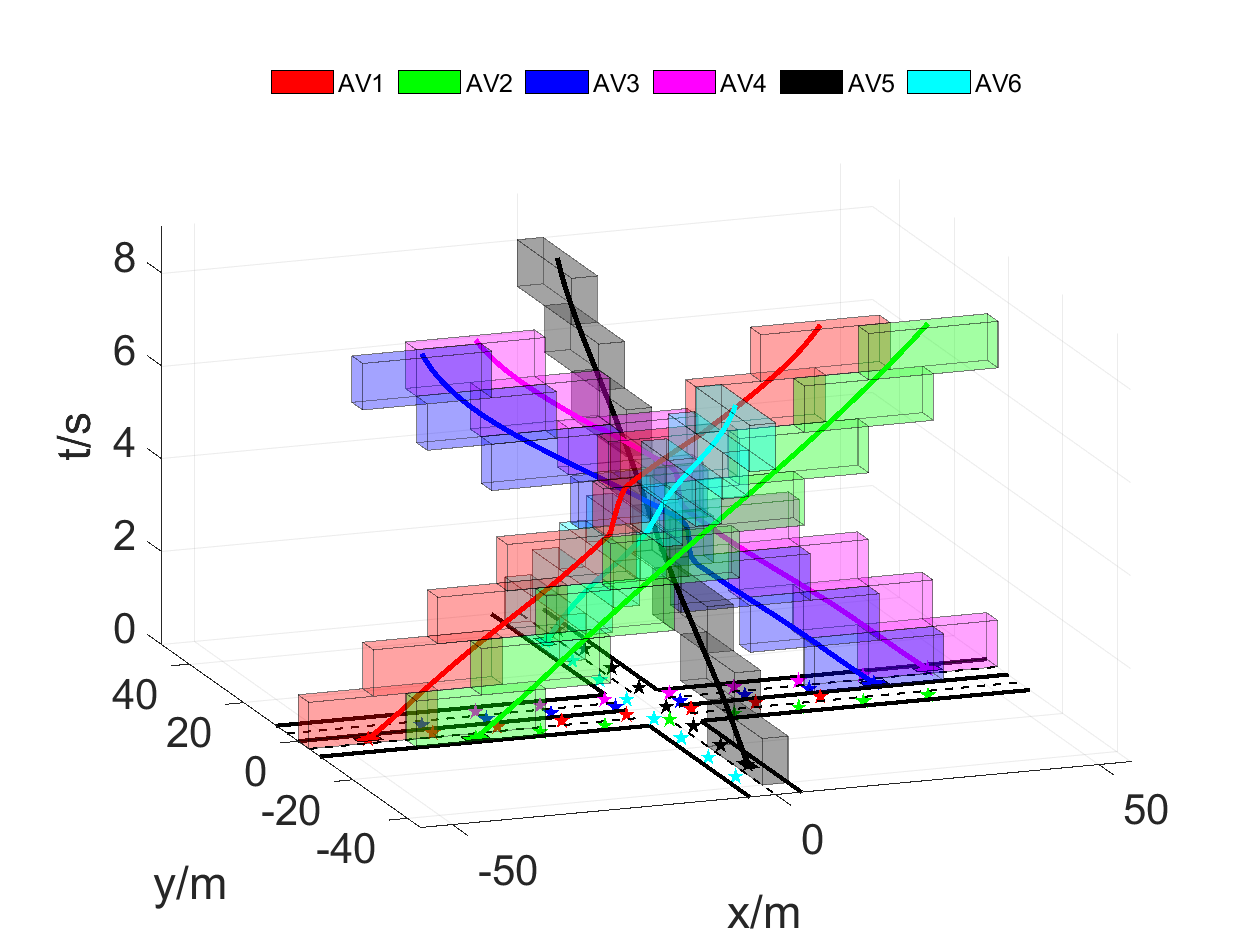}
  \label{fig.stc_noauto_n1}}
  \quad
  \subfloat[]{
  \includegraphics[width=3.3in]{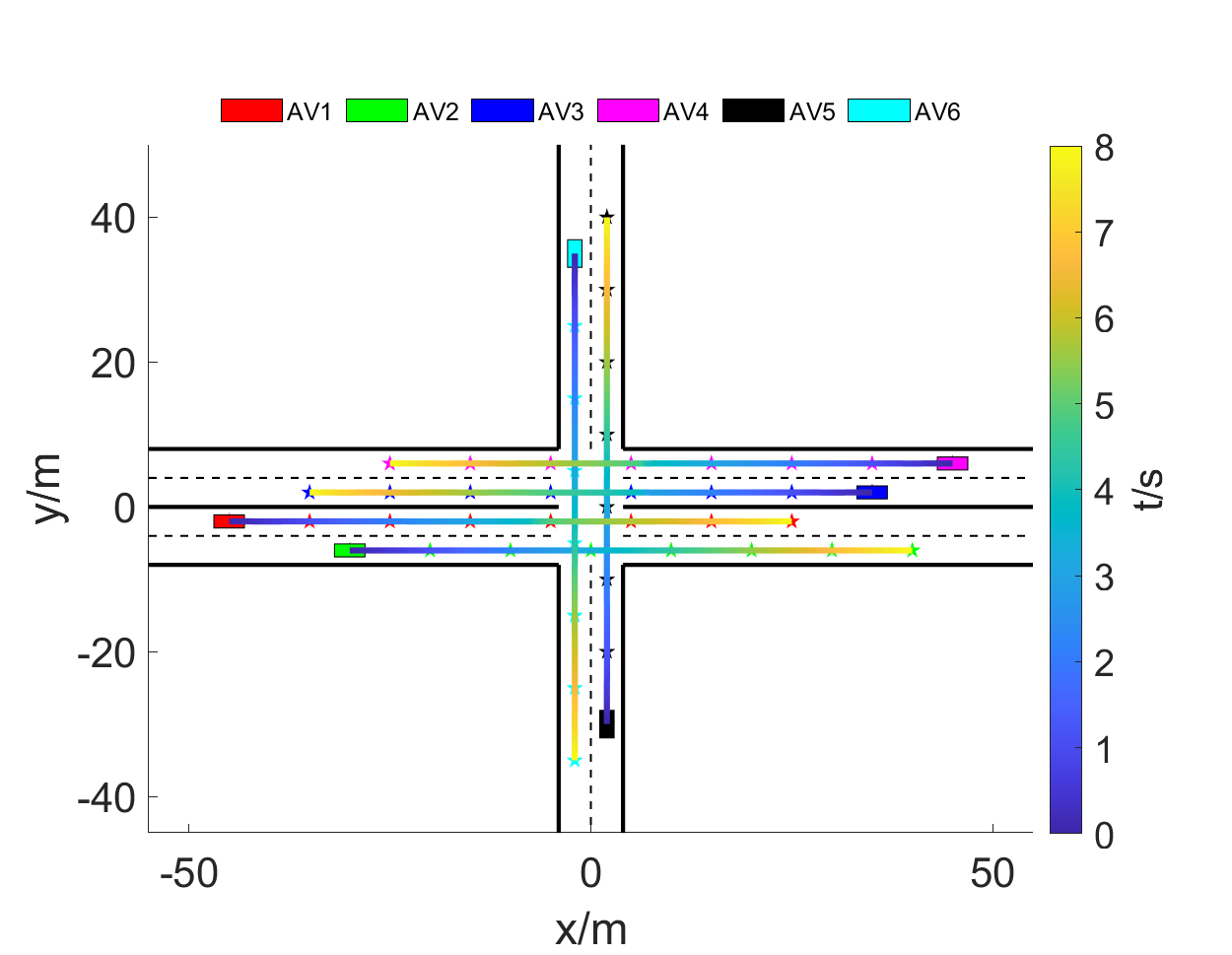}
  \label{fig.tra_noauto_n1}
  }
  \caption{Simulation results of the STC method \cite{Zhang2023} in the unsignalized intersection scenario: (a) safety corridors; (b) trajectories.}
  \label{fig.stc_tra_noauto_n1}
\end{figure*}

\begin{table}[htp] 
  \caption{Initial and target positions of AVs in the unsignalized intersection scenario}
  \label{positions_n1}
  \centering  
    \begin{tabular}{|c|c|c|c|} 
    \hline
          Vehicles & Initial position($m$) & Target position($m$) \\
    \hline 
    AV1 & (-45, -2)  & (25,-2) \\
    \hline
    AV2 & (-30, -6)  & (40,-6) \\
    \hline
    AV3 & (35,2)  & (-35,2) \\
    \hline
    AV4 & (45,6)  & (-25,6) \\
    \hline
    AV5 & (2,-30) & (2,40)  \\
    \hline
    AV6 & (-2,35) & (-2,-35) \\
    \hline
  \end{tabular}
\end{table}

All simulations are implemented in C++ for computational efficiency, while visualization and result analysis are performed in MATLAB. All experiments are conducted on a computer running Ubuntu 18.04 with an Intel Core i7-10850H CPU at 2.70 GHz and 8 GB RAM. The Gurobi solver \cite{Gurobi2023} is used to solve \eqref{MIQP} for constructing the V-STC, and the Ipopt solver \cite{Wächter2006} is used to solve \eqref{NLP} for generating the final trajectory inside each AV's V-STC. A dynamic demonstration of the simulated trajectories is available at the following link\footnote{\url{https://github.com/BIT-MVTP/V-STC/issues/1}}.

\paragraph{Unsignalized Intersection Scenario}
In this scenario, six AVs travel on different lanes of an unsignalized intersection. AVs 1-4 move horizontally, while AVs 5 and 6 move longitudinally. As the AVs enter the shared intersection area from orthogonal directions, their routes intersect within the conflict region, leading to multiple interaction and collision risks. The lane width is set to 4 $m$. The initial and target positions of all AVs are listed in Table \ref{positions_n1}, with an initial speed of 10 $m/s$ for each AV. The positions are defined in a Cartesian coordinate system, where $(0, 0)$ denotes the intersection center. The simulation results of the two methods in this scenario are shown in Fig. \ref{fig.stc_tra_auto_n1} and Fig. \ref{fig.stc_tra_noauto_n1}.

\begin{table}[htp] 
  \caption{Initial and target positions of AVs in the multi-lane lane-change scenario}
  \label{positions_n2}
  \centering  
    \begin{tabular}{|c|c|c|c|}
    \hline
          Vehicles & Initial position($m$) & Target position($m$) \\
    \hline 
    AV1 & (5, 2)  & (75, 6) \\
    \hline
    AV2 & (40, 6)  & (110, 10) \\
    \hline
    AV3 & (0, 10)  & (70, 6) \\
    \hline
    AV4 & (25,14)  & (95,10) \\
    \hline
  \end{tabular}
\end{table}

\begin{figure*}[htp]
  \centering
  \subfloat[]{
  \includegraphics[width=3.3in]{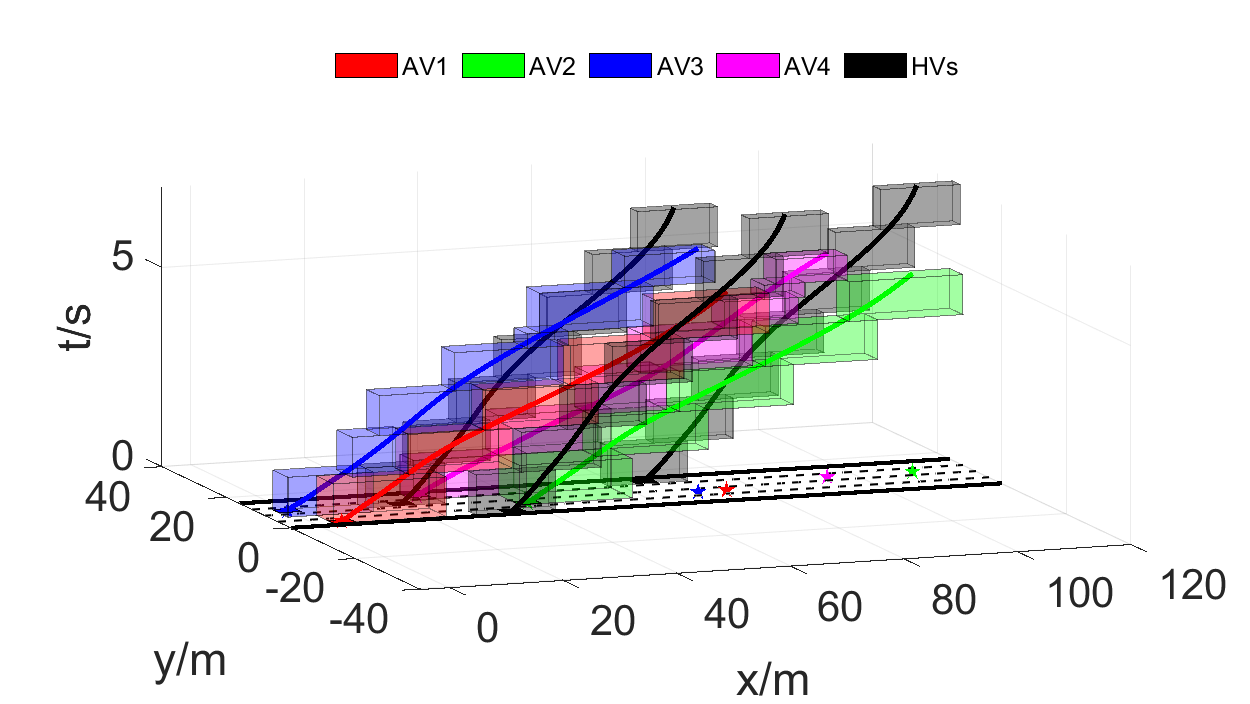}
  \label{fig.stc_auto_n2}}
  \quad
  \subfloat[]{
  \includegraphics[width=3.3in]{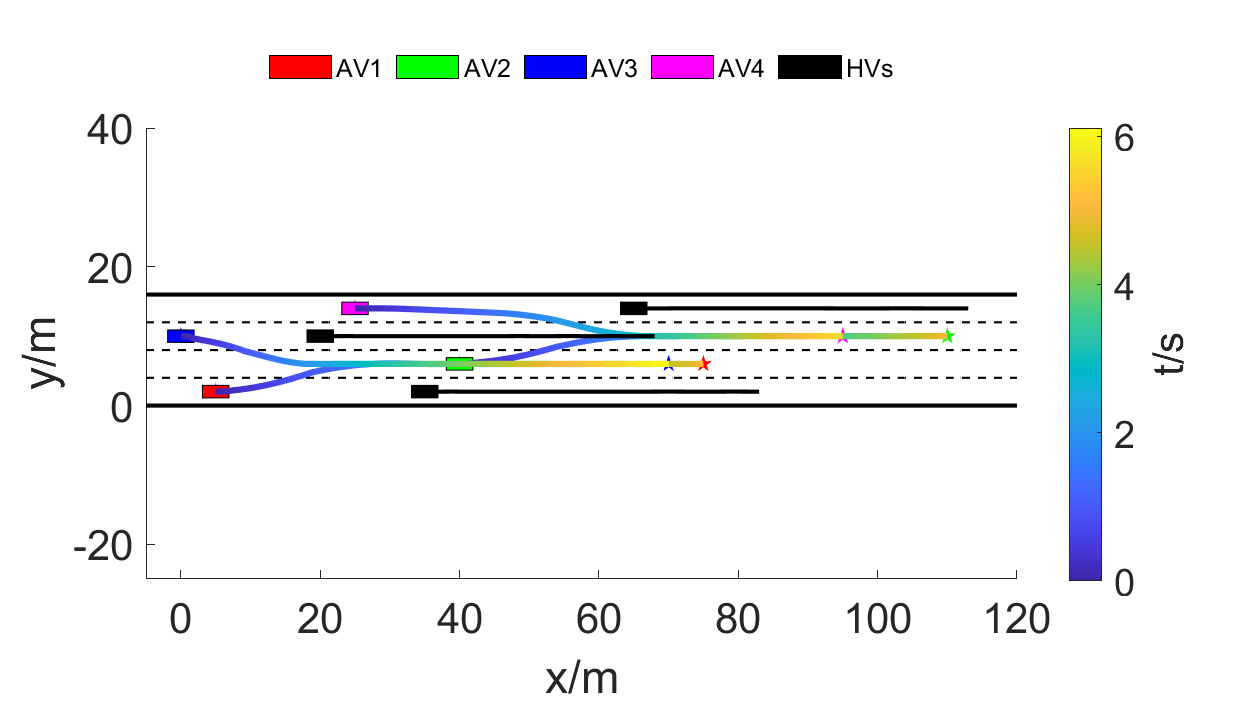}
  \label{fig.tra_auto_n2}
  }
  \caption{Simulation results of the proposed V-STC method in the lane-change scenario: (a) safety corridors; (b) trajectories.}
  \label{fig.stc_tra_auto_n2}
\end{figure*}

\begin{figure*}[hbp]
  \centering
  \subfloat[]{
  \includegraphics[width=3.3in]{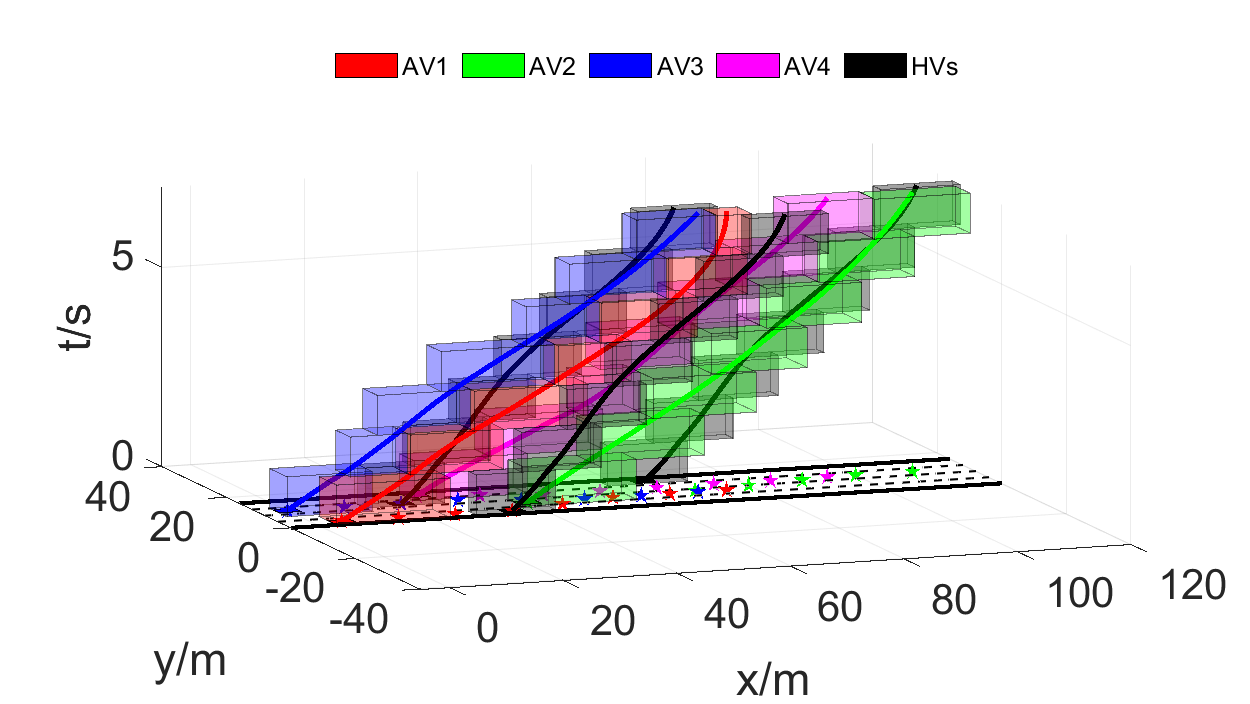}
  \label{fig.stc_noauto_n2}}
  \quad
  \subfloat[]{
  \includegraphics[width=3.3in]{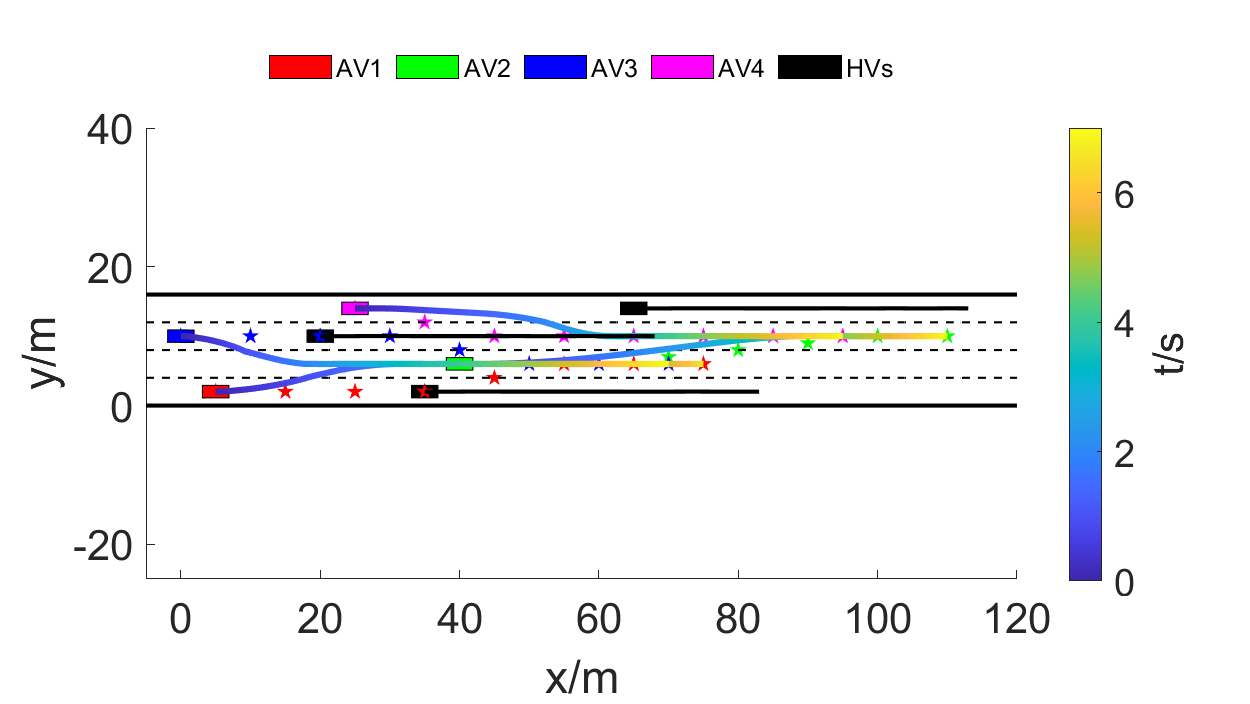}
  \label{fig.tra_noauto_n2}
  }
  \caption{Simulation results of the STC method \cite{Zhang2023} in the lane-change scenario: (a) safety corridors; (b) trajectories.}
  \label{fig.stc_tra_noauto_n2}
\end{figure*}

\paragraph{Multi-Lane Lane-Change Scenario}
In this scenario, four AVs and three HVs travel on a straight four-lane road with a lane width of 4 $m$. The vehicle positions are defined in a Cartesian coordinate system, where the $x$-axis represents the longitudinal direction and the $y$-axis represents the lateral direction of the road. The lane centers are located at $y$ = 2, 6, 10, and 14 $m$. The HVs are modeled as spatio-temporal corridor cubes with known occupancy. The initial and target positions of the AVs are listed in Table \ref{positions_n2}, with an initial speed of 10 $m/s$ for each AV. The AVs coordinate their longitudinal motion and lane-change maneuvers to avoid collisions while reaching their target positions and lanes. The simulation results of the compared methods are shown in Fig. \ref{fig.stc_tra_auto_n2} and Fig. \ref{fig.stc_tra_noauto_n2}.

\begin{figure*}[hbp]
  \centering
  \subfloat[]{
  \includegraphics[width=3.3in]{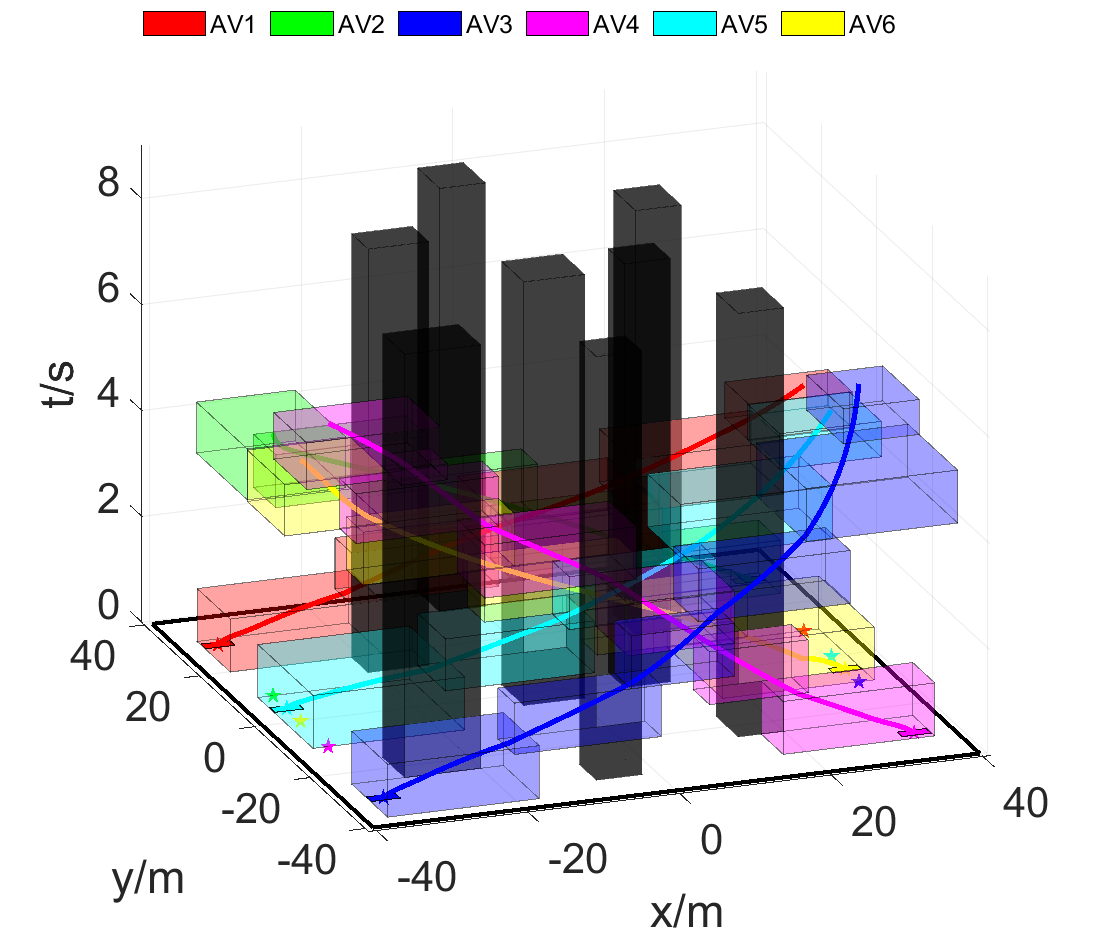}
  \label{fig.stc_auto_1}}
  \quad
  \subfloat[]{
  \includegraphics[width=3.3in]{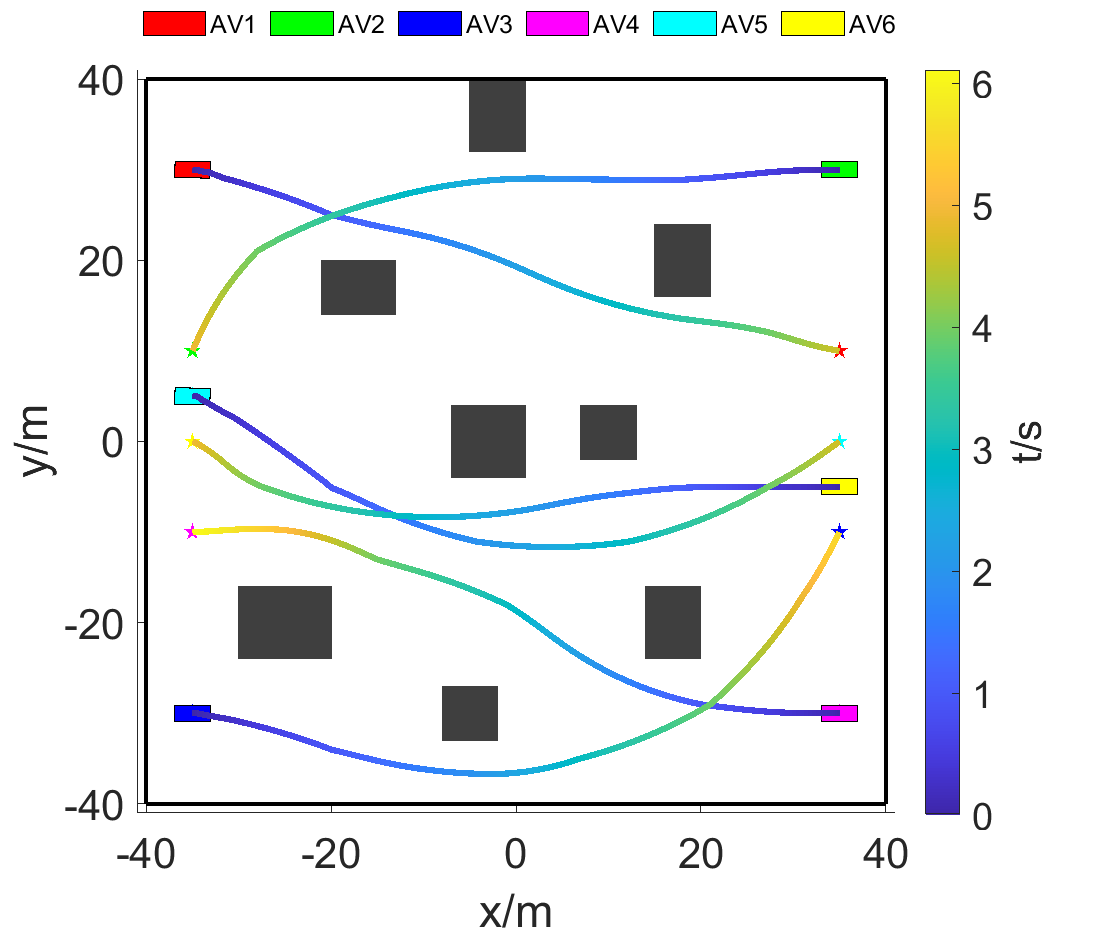}
  \label{fig.tra_auto_1}
  }
  \caption{Simulation results of the proposed V-STC method in the unstructured environment scenario: (a) safety corridors; (b) trajectories.}
  \label{fig.stc_tra_auto_1}
\end{figure*}

\begin{figure*}[hbp]
  \centering
  \subfloat[]{
  \includegraphics[width=3.3in]{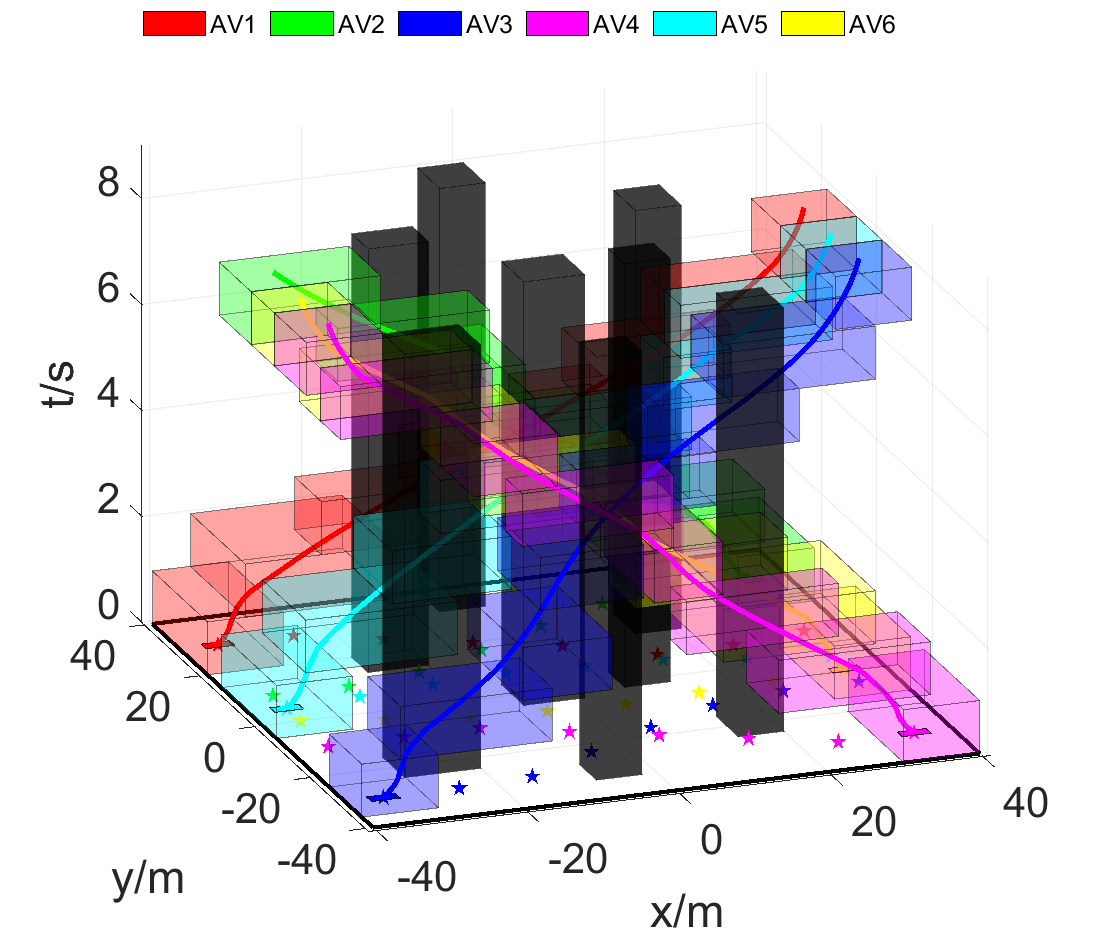}
  \label{fig.stc_noauto_1}}
  \quad
  \subfloat[]{
  \includegraphics[width=3.3in]{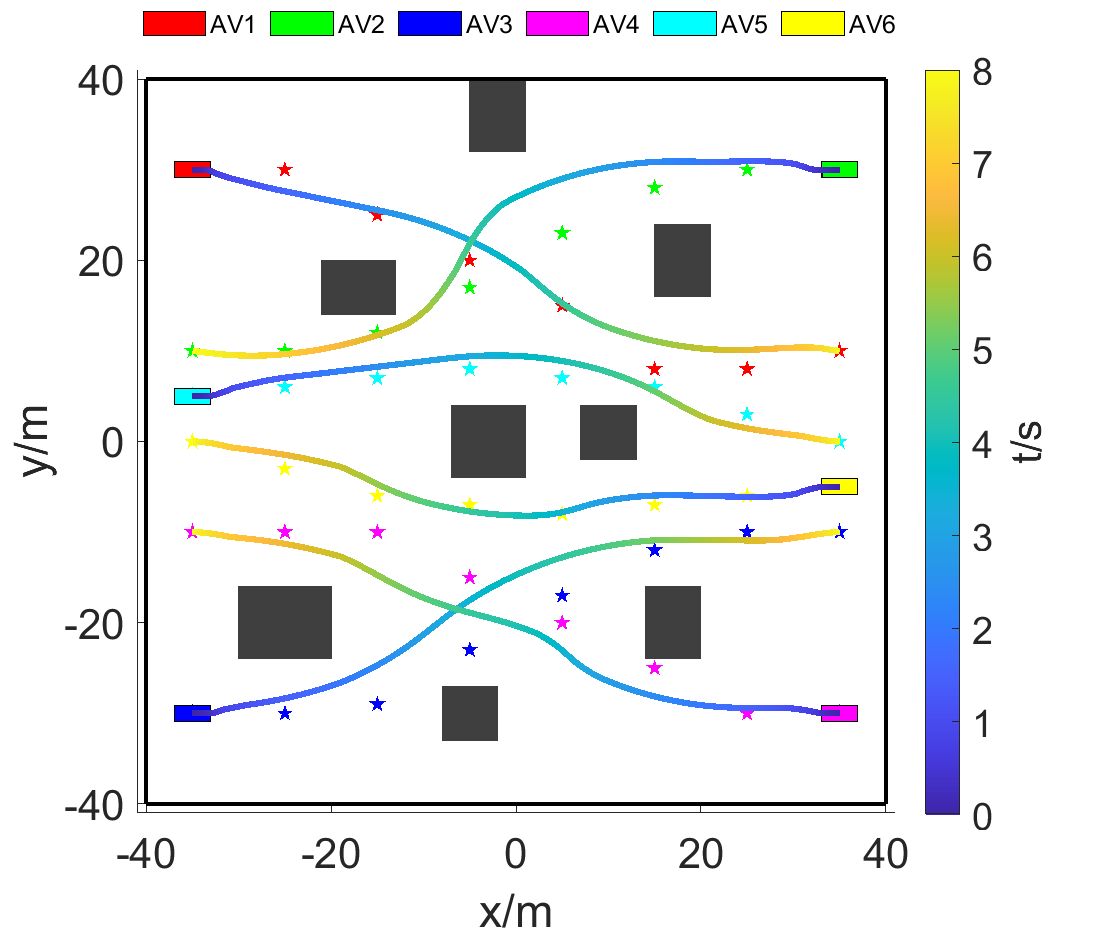}
  \label{fig.tra_noauto_1}
  }
  \caption{Simulation results of the STC method \cite{Zhang2023} in the unstructured environment scenario: (a) safety corridors; (b) trajectories.}
  \label{fig.stc_tra_noauto_1}
\end{figure*}

\paragraph{Unstructured Environment Scenario}
In this scenario, six AVs move in an unstructured environment defined over an 80 $m$ $\times$ 80 $m$ area, where the center of the map is set to $(0,0)$ in the Cartesian coordinate system. The environment contains eight static obstacles with centers located at $(-25,-20)$, $(17,-20)$, $(18,20)$, $(-3,0)$, $(10,1)$, $(-2,36)$, $(-5,-30)$, and $(-17,17)$, respectively. The corresponding length--width pairs are $(10,8)$, $(6,8)$, $(6,8)$, $(8,8)$, $(6,6)$, $(6,8)$, $(6,6)$, and $(8,6)$. The initial and target positions of all AVs are listed in Table \ref{positions_1}, and the initial speed of each AV is set to 10 $m/s$. AVs 1, 3, and 5 move from left to right, while AVs 2, 4, and 6 move from right to left. Static obstacles and intersecting routes introduce potential collision risks that must be resolved through trajectory planning. The simulation results of the compared methods are shown in Fig. \ref{fig.stc_tra_auto_1} and Fig. \ref{fig.stc_tra_noauto_1}.

\begin{table}[htp] 
  \caption{Initial and target positions of AVs in the unstructured environment scenario}
  \label{positions_1}
  \centering  
    \begin{tabular}{|c|c|c|c|}
    \hline
          Vehicles & Initial position($m$) & Target position($m$) \\
    \hline 
    AV1 & (-35, 30)  & (35, 10) \\
    \hline
    AV2 & (35, 30)  & (-35, 10) \\
    \hline
    AV3 & (-35,-30)  & (35, -10) \\
    \hline
    AV4 & (35,-30)  & (-35, -10) \\
    \hline
    AV5 & (-35, 5)  & (35, 0) \\
    \hline
    AV6 & (35, -5)  & (-35, 0) \\
    \hline
  \end{tabular}
\end{table}

\paragraph{Results and Discussion}
The simulation results in the three scenarios demonstrate clear performance differences between the proposed V-STC method and the baseline STC method \cite{Zhang2023}. While both methods generate feasible collision-free trajectories in all scenarios, V-STC consistently achieves shorter trajectory completion times, indicating higher efficiency in multi-vehicle coordinated maneuvers.

For the unsignalized intersection scenario, as shown in Fig. \ref{fig.stc_tra_auto_n1}, the corridors of the lateral AVs 1-4 utilize the gaps between the corridors of the longitudinal AVs 5 and 6 to traverse the intersection smoothly. This allows the vehicles to approach and pass through the conflict region without prolonged stopping or waiting. In contrast, under the STC method, the corridors of AVs 5 and 6 occupy the conflict region earlier, as shown in Fig. \ref{fig.stc_tra_noauto_n1}. As a result, when the lateral AVs 1, 3, and 4 attempt to traverse the intersection, their corridor cubes at 5 $s$ are constrained by the spatio-temporal occupancy of AVs 5 and 6, leading to more pronounced deceleration near the intersection. The trajectory results further show that the maximum completion time is approximately 5 $s$ under V-STC, compared with 8 $s$ under STC.

For the multi-lane lane-change scenario, the main difference between the two methods lies in the timing of lane-change completion and its influence on the following vehicles. As shown in Fig. \ref{fig.stc_tra_auto_n2}, under the V-STC method, AVs 1-4 complete their lane changes shortly after start, resulting in more compact spatio-temporal corridors. In contrast, under the STC method, the corridors of vehicles at the same time step constrain each other during the lane-change process, as shown in Fig. \ref{fig.stc_tra_noauto_n2}. This delays the lane-change completion of AVs 2 and 4, which subsequently affects the longitudinal motion of the following AVs 1 and 3.

For the unstructured environment scenario, the main advantage of V-STC lies in its ability to construct corridors without relying on a predefined waypoint distribution. As shown in Fig. \ref{fig.stc_tra_auto_1}, under V-STC, AV2 detours through a wider passage near the map boundary, thereby avoiding direct interaction with AV1 in a narrow area. As a result, both AVs 1 and 2 maintain nearly straight trajectories and reach their target positions at relatively high speeds. Similarly, AV3 selects a wide region near the boundary, reducing interactions with other AVs in the central area. In contrast, under the STC method, as shown in Fig. \ref{fig.stc_tra_noauto_1}, the spatio-temporal corridors of the AVs are more tightly constrained by the predefined waypoint distribution, forcing AVs to pass through narrow areas at similar times. For example, AVs 1 and 2, as well as AVs 3 and 4, converge near the center of the map, which leads to deceleration during passage.

\begin{table*}[htp] 
  \caption{Comparison of Trajectory Durations in Different Simulation Scenarios}
  \label{Time}
  \centering  
  \begin{tabular}{|c|c|c|c|  c|c|c|c|c|c| }
    \hline
    \multirow{2}*{Scenario} & \multirow{2}*{Obstacles}  &  \multirow{2}*{$N_{AV}$} & \multirow{2}*{$t_{STC}$(s) \cite{Zhang2023}} &  \multicolumn{6}{c|}{$t_{V-STC}$(s)} \\ 
    \cline{5-10}
            &    &    &   & AV1  & AV2 & AV3  & AV4  & AV5 & AV6  \\
    \hline 
    (a) & {$\times$ }       & 6  & 8.00 & 5.93 & 4.57 & 5.57 & 4.97 & 4.56 & 5.13 \\
    \hline
    (b) & {$\checkmark$}    & 4  & 7.00 & 4.99 & 4.97 & 6.11 & 5.61 & - & - \\
    \hline
    (c) & {$\checkmark$}    & 6  & 8.00 & 4.64 & 4.96 & 5.63 & 6.11 & 4.65 & 4.94 \\
    \hline
  \end{tabular}
\end{table*}

Table \ref{Time} summarizes the trajectory durations of each AV generated by the V-STC and STC methods across different scenarios. In all cases, the trajectory durations $t_{V\text{-}STC}$ are shorter than the corresponding $t_{STC}$, indicating higher efficiency of the proposed method. By avoiding unnecessary waypoint pre-setting and temporal synchronization, V-STC effectively reduces the overall travel time.

\section{Conclusion}
\label{Conclusion}
 This paper presented a V-STC framework for coordinated multi-vehicle trajectory planning. The proposed approach jointly determines the spatial boundaries and temporal durations of corridor cubes, enabling the generation of collision-free and time-efficient corridors without relying on preassigned waypoints. By introducing the time step of each corridor cube as an optimization variable, the method adaptively adjusts the temporal allocation along each AV's path, thereby reducing the overall occupancy time in interaction regions. Based on the constructed V-STC, the multi-vehicle planning problem is decomposed into a set of single-vehicle trajectory optimization problems, each solved within its own corridor. The resulting trajectories satisfy AV dynamics constraints and remain safely inside the constructed V-STC. Simulation results in structured and unstructured environments demonstrate that the proposed method consistently reduces traversal time while ensuring safe and smooth multi-vehicle coordination. Future work will focus on extending the V-STC-based distributed multi-vehicle coordination framework to enhance its adaptability in high-density traffic flow environments, which will serve as a critical research direction for improving the framework's robustness.





\section{Biography Section}
\begin{IEEEbiography}[{\includegraphics[width=1in,height=1.25in,clip,keepaspectratio]{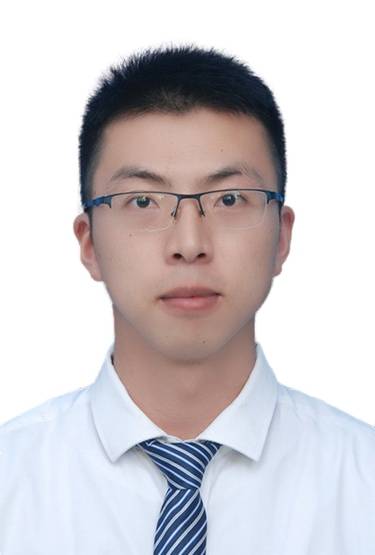}}]
  {Pengfei Liu} received the B.E. degree in Electrical Engineering and Automation from Hefei University of Technology, Hefei, China in 2020, and the M. S. degree in Aeronautical Engineering from Beihang University, Beijing, China in 2023. He is currently pursuing the Ph.D. degree in control engineering with the School of Automation, Beijing Institute of Technology, Beijing, China. His research interests include cooperative trajectory planning of multi-robot systems, nonlinear dynamics and robot control.
 \end{IEEEbiography}

 \vspace{11pt}
 
 \begin{IEEEbiography}[{\includegraphics[width=1in,height=1.25in,clip,keepaspectratio]{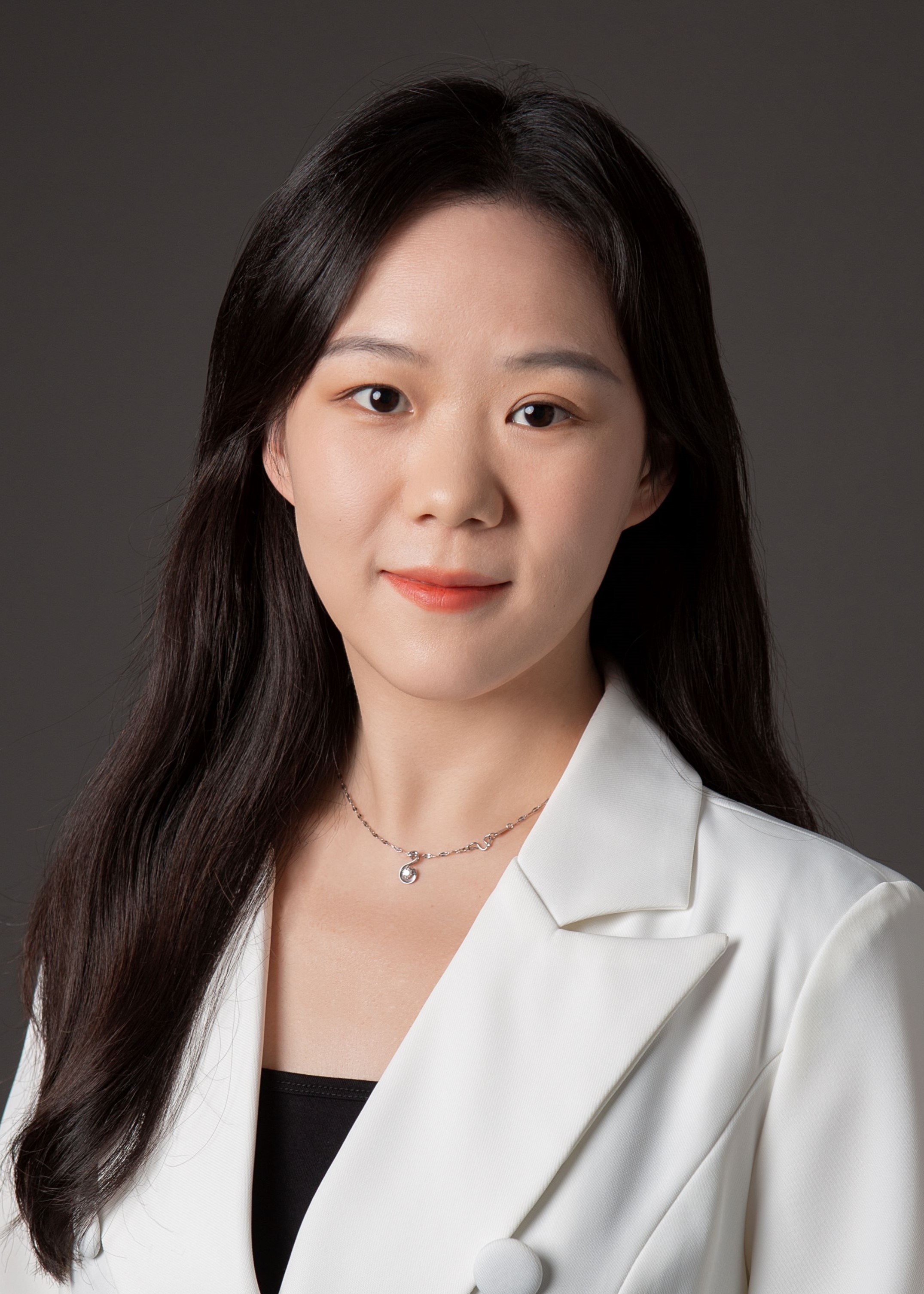}}]
   {Jialing Zhou} (Senior Member, IEEE) received the Ph.D. degree in mechanical systems and control from Peking University, Beijing, China, in 2017. 
   
   She is currently a Professor with the School of Interdisciplinary Science, Beijing Institute of Technology, Beijing, China. Her research interests include motion planning, distributed control and optimization, reinforcement learning, and networked games.
 
   Dr. Zhou was selected as a Young Top-Notch Talent under the National Ten Thousand Talents Program and was the recipient of the Young Elite Scientist Sponsorship Program from the China Association for Science and Technology. She is an Associate Editor for \textit{Asian Journal of Control}.
 
  \end{IEEEbiography}
  
  \vspace{11pt}
 
  \begin{IEEEbiography}[{\includegraphics[width=1in,height=1.25in,clip,keepaspectratio]{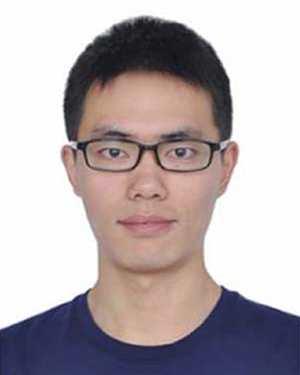}}]
     {Yuezu Lv} (Senior Member, IEEE) received the B.S. degree in engineering mechanism and the Ph.D. degree in mechanical systems and control from the College of Engineering, Peking University, Beijing, China, in 2013 and 2018, respectively. 
     
     He is currently a Professor with the School of Artificial Intelligence, Beijing Institute of Technology, Beijing. His research interests include cooperative control of multiagent systems, adaptive control, robust control of uncertain systems, and distributed resilient control. 
     
     Dr. Lv was the recipient of the 2021 APNNS Young Researcher Award by Asia Pacific Neural Network Society. He is an Associate Editor of \textit{IEEE SMC Magazine}, and a Young Editorial Board Member of \textit{International Journal of Dynamics and Control}.
   \end{IEEEbiography}
 
   \vspace{11pt}
 
    \begin{IEEEbiography}[{\includegraphics[width=1in,height=1.25in,clip,keepaspectratio]{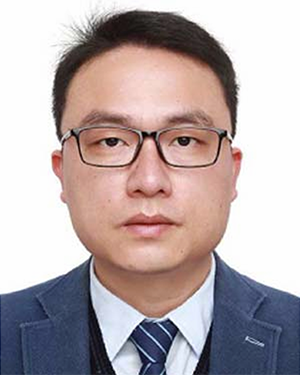}}]
     {Guanghui Wen} (Senior Member, IEEE) received the Ph.D. degree in mechanical systems and control from Peking University, Beijing, China, in 2012.
     
     Currently, he is an Endowed Chair Professor and the Vice Dean of the School of Automation, Southeast University, Nanjing, China. His current research interests include autonomous intelligent systems, complex networked systems, distributed control and optimization, resilient control, and distributed reinforcement learning.
     
     Prof. Wen was the recipient of the National Science Fund for Distinguished Young Scholars, the China Youth Science and Technology Award, the Australian Research Council Discovery Early Career Researcher Award, and the Asia Pacific Neural Network Society Young Researcher Award. He currently serves as a Technical Editor of the \textit{IEEE/ASME Transactions on Mechatronics} and an Associate Editor of the \textit{IEEE Transactions on Control of Network Systems}, the \textit{IEEE Transactions on Industrial Informatics}, the \textit{IEEE Transactions on Neural Networks and Learning Systems}, the \textit{IEEE Transactions on Intelligent Vehicles}, the \textit{IEEE Transactions on Fuzzy Systems}, the \textit{IEEE Transactions on Systems, Man and Cybernetics: Systems}, the \textit{IEEE Open Journal of the Industrial Electronics Society}, and the \textit{Asian Journal of Control}. Prof. Wen has been named a Highly Cited Researcher by Clarivate Analytics since 2018. He is an IET Fellow.
   \end{IEEEbiography}
 
   \vspace{11pt}
 
   \begin{IEEEbiography}[{\includegraphics[width=1in,height=1.25in,clip,keepaspectratio]{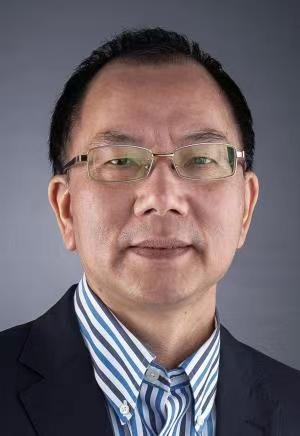}}]
     {Tingwen Huang} (Fellow, IEEE) received the B.S. degree in mathematics from Southwest Normal University, Chongqing, China, in 1990, the M.S. degree in applied mathematics from Sichuan University, Chengdu, China, in 1993, and the Ph.D. degree in mathematics from Texas A$\&$M University, College Station, TX, USA, in 2002.
 
     He was a Lecturer with Jiangsu University, Zhenjiang, China, from 1994 to 1998, and a Visiting Assistant Professor with Texas A$\&$M University in 2003. From 2003 to 2009, he was an Assistant Professor, from 2009 to 2013, an Associate Professor, and since 2013, has been a Professor with Texas A$\&$M University at Qatar, Doha, Qatar. He has been a Professor with the Faculty of Computer Science and Control Engineering, Shenzhen University of Advanced Technology, Shenzhen, China, since 2024. His research interests include neural networks, complex networks, chaos and dynamics of systems, and operator semi-groups and their applications.
   \end{IEEEbiography}

\vfill

\end{document}